\documentclass{article}

\usepackage{microtype}
\usepackage{graphicx}
\usepackage{subcaption}
\usepackage{enumitem}
\usepackage{booktabs} 

\usepackage{hyperref}

\usepackage[accepted]{icml2025}

\usepackage{amsmath}
\usepackage{amssymb}
\usepackage{mathtools}
\usepackage{amsthm}
\usepackage{multirow}
\usepackage[capitalize,noabbrev]{cleveref}

\theoremstyle{plain}
\newtheorem{theorem}{Theorem}[section]

\theoremstyle{definition}
\newtheorem{definition}[theorem]{Definition}

\theoremstyle{remark}

\usepackage[textsize=tiny]{todonotes}

\icmltitlerunning{Submission and Formatting Instructions for ICML 2025}

\begin{document}

\twocolumn[
\icmltitle{Exploring Implicit Visual Misunderstandings in \\
Multimodal Large Language Models through Attention Analysis}



\icmlsetsymbol{equal}{*}

\begin{icmlauthorlist}
\icmlauthor{Pengfei Wang}{sch,comp}
\icmlauthor{Guohai Xu}{comp}
\icmlauthor{Weinong Wang}{comp}
\icmlauthor{Junjie Yang}{comp}
\icmlauthor{Jie Lou}{comp}
\icmlauthor{Yunhua Xue}{sch}
\end{icmlauthorlist}

\icmlaffiliation{sch}{School of Mathematical Sciences, Nankai University, Tianjin, China}
\icmlaffiliation{comp}{Xiaohongshu Inc, Shanghai, China}

\icmlcorrespondingauthor{Guohai Xu}{guohai.explorer@gmail.com}

\icmlkeywords{Implicit Visual Misunderstandings, Attention Accuracy}

\vskip 0.3in
]



\printAffiliationsAndNotice{}  

\begin{abstract}
Recent advancements have enhanced the capability of Multimodal Large Language Models (MLLMs) to comprehend multi-image information.
However, existing benchmarks primarily evaluate answer correctness, overlooking whether models genuinely comprehend the visual input.
To address this, we define implicit visual misunderstanding (IVM), where MLLMs provide correct answers without fully comprehending the visual input.
Through our analysis, we decouple the visual and textual modalities within the causal attention module, revealing that attention distribution increasingly converges on the image associated with the correct answer as the network layers deepen.
This insight leads to the introduction of a scale-agnostic metric, \textit{attention accuracy}, and a novel benchmark for quantifying IVMs.
Attention accuracy directly evaluates the model's visual understanding via internal mechanisms, remaining robust to positional biases for more reliable assessments.
Furthermore, we extend our approach to finer granularities and demonstrate its effectiveness in unimodal scenarios, underscoring its versatility and generalizability.
\end{abstract}
\section{Introduction}
\label{sec:intro}
\begin{figure}[t]
    \centering
    \includegraphics[width=\linewidth]{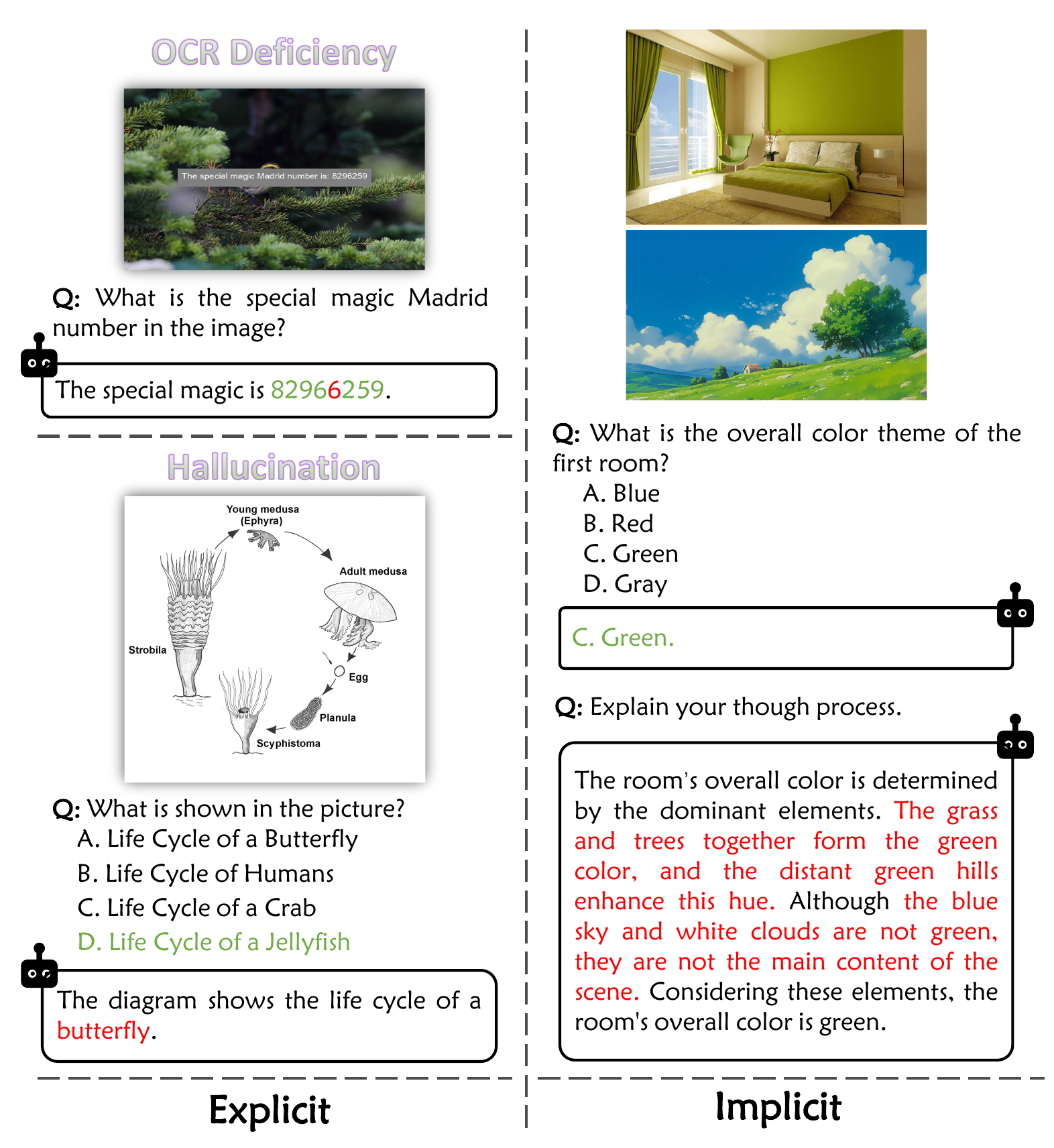}
    \vspace{-15pt}
    \caption{
    \textbf{Left}: Example of explicit visual misunderstandings: OCR deficiency and hallucination.
    \textbf{Right}: Example of implicit visual misunderstandings: the model provides a correct answer but actually describes the second image (while the question pertains to the content of the first image).
    }
    \vspace{-5pt}
    \label{fig:1_examples}
\end{figure}
\begin{figure*}[t]
    \centering
    \includegraphics[width=\linewidth]{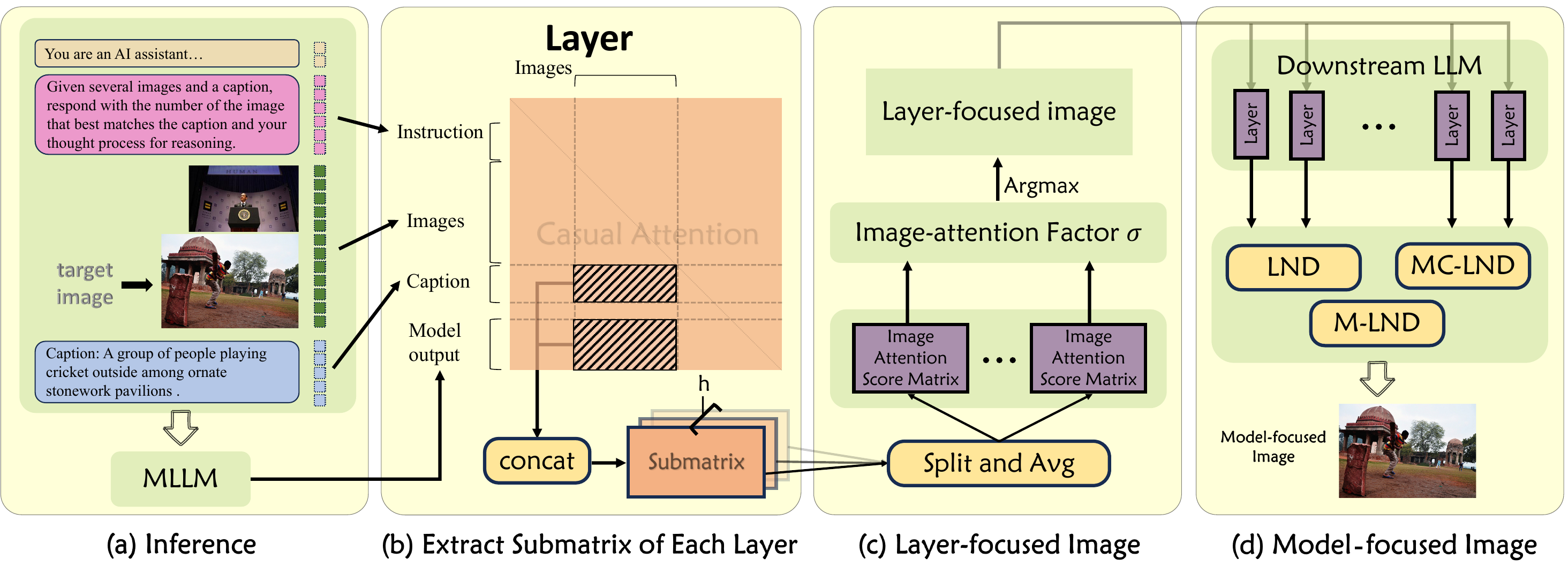}
    \vspace{-15pt}
    \caption{\textbf{Overview of the Approach for Identifying Model-focused Image.}~~
    (a) MLLM completes a caption-matching task;
    (b) the attention submatrix for multimodal interactions is extracted;
    (c) for each layer, attention factor values for each image are calculated, allowing identification of the layer-focused image;
    (d) finally, the model-focus image is determined using the three metrics.
    }
    \vspace{-6pt}
    \label{fig:2_attn}
\end{figure*}
MLLMs \cite{qwen2vl,internvl2,gpt-4o} have demonstrated remarkable performance in handling multi-image tasks.
However, as the number of images increases, limitations in training data and imbalances in training methodologies \cite{MLLM_survey} lead to a higher incidence of visual misunderstandings.
We distinguish between two forms of visual misunderstanding: explicit and implicit, as shown in \cref{fig:1_examples}.
\textbf{Explicit visual misunderstandings} (EVMs) occur when models provide incorrect answers, making their deficiencies in visual abilities easily identifiable—for instance, errors stemming from OCR deficiencies or hallucinations.
\textbf{Implicit visual misunderstandings} refer to cases where models deliver correct answers despite misinterpreting or misunderstanding the corresponding visual content.
Ideally, a lower incidence of EVMs would signal stronger visual capabilities in MLLMs, but the presence of IVMs introduces complexity to this relationship.

Numerous tasks have been proposed to evaluate the visual understanding capabilities of MLLMs.
MMVP \cite{MMVP} uses ``CLIP-blind pairs'' to delve the failures of the visual encoder \cite{EvaCLIP,SigLIP} in models.
Some studies \cite{POPE,MMHAL-BENCH,object_hallucination} specifically focus on explicit hallucinations.
Other benchmarks assess models across various aspects, such as the necessity of visual information \cite{MMIU,MMStar} and visual illusions \cite{HallusionBench}.
Nevertheless, all these methods share a \textbf{common limitation}:
they focus solely on the correctness of MLLMs' answers (i.e., EVMs), without considering whether the models truly understand the target visual content.

The occurrence of IVMs is also closely tied to the design of existing benchmarks and the structure of the MLLMs.
Current multi-image benchmarks mainly adopt multiple-choice questions \cite{MileBench,Seed}, which may inadvertently allow models to guess correct answers without fully analyzing the visual input \cite{ScienceQA}.
Additionally, MLLMs often leverage extensive prior knowledge stored in downstream LLMs, enabling them to provide seemingly accurate responses by relying on memorized patterns or textual correlations from their training data \cite{Visualbert,MMStar}.
Therefore, these models may bypass the need for genuine visual understanding, masking their actual limitations in processing visual information.

In this work, we perform a quantitative analysis of IVMs in MLLMs, effectively overcoming the challenge of their inability to be explicitly evaluated.
As a first step, we decouple the visual and textual modalities within the causal attention module.
Our findings reveals an intriguing pattern: in multi-image scenarios, although different attention heads focus on various visual regions, their aggregated scores consistently concentrate on the target image—the one linked to the correct answer.
This phenomenon is especially prominent in well-trained, large-scale models \cite{qwen2vl, llava-ov}.
Inspired by this, we introduce the \textbf{S}ingle-\textbf{T}arget \textbf{M}ultimodal \textbf{E}valuation (\textbf{STME}) benchmark, which incorporates two levels of difficulty and covers diverse tasks such as caption matching \cite{Flickr30k} and OCR recognition \cite{coco2014}.
Using STME, we define attention accuracy as a metric to quantify the extent of IVMs.

Experiments proves that attention accuracy offers a more comprehensive evaluation of MLLMs' capabilities from a visual perspective, remaining unaffected by positional biases in the images.
It serves as an equivariant measure, which means it can reliably assess IVMs across different model series, architectures, and scales.
This consistency allows for uniform evaluation of models across visual tasks with varying categories and levels of difficulty.
Finally, we extend the approach to a finer-grained token level and apply them to scenarios involving single-modal interactions.
This further enhances the metrics’ versatility and offers deeper insights into the interactions between modalities within MLLMs.

Overall, our contributions are summarized as follows:
\begin{itemize}[noitemsep,leftmargin=*]
\vspace{-8px}
\item Our findings reveal that as the layers deepen, attention converges onto a specific image, Based on this, we propose a quantitative metric to measure the attention distribution across all images within any layer of the MLLMs.
\item We design the STME benchmark, a novel dataset tailored for single-target visual tasks across diverse domains, providing a foundation for evaluating IVMs in MLLMs.
\item We introduce attention accuracy to characterize IVMs in MLLMs, enabling consistent evaluation across models of various series, scales, training stages, and architectures, while also being the first to assess model capabilities from their internal mechanisms.
\end{itemize}
\begin{figure*}[ht]
    \centering
    \begin{subfigure}{0.45\linewidth}
        \centering
        \includegraphics[width=\linewidth]{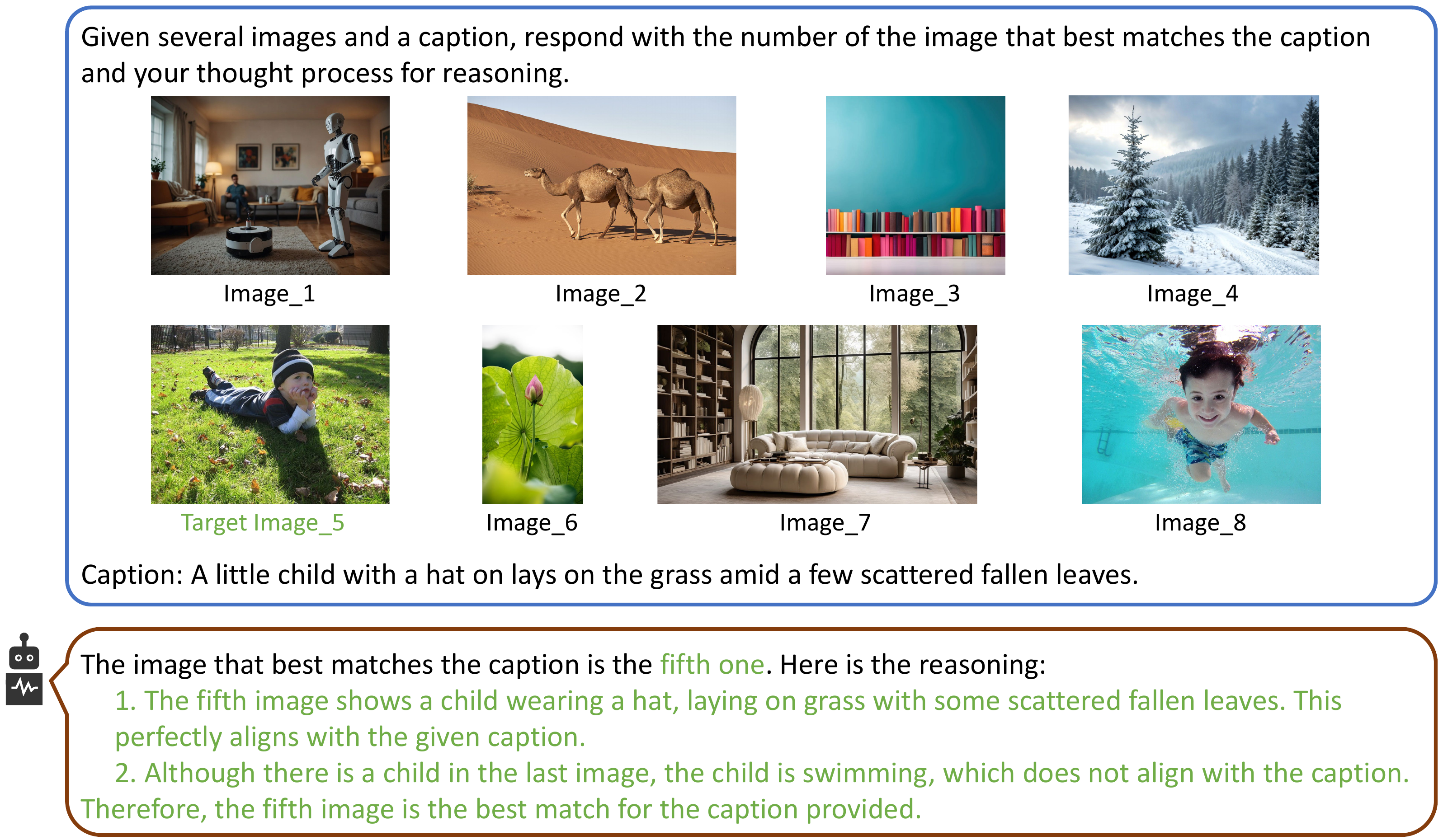}
        \label{fig:3_1_1}
    \end{subfigure}
    \hfill
    \vspace{-18px}
    \begin{subfigure}{0.52\linewidth}
        \centering
        \includegraphics[width=\linewidth]{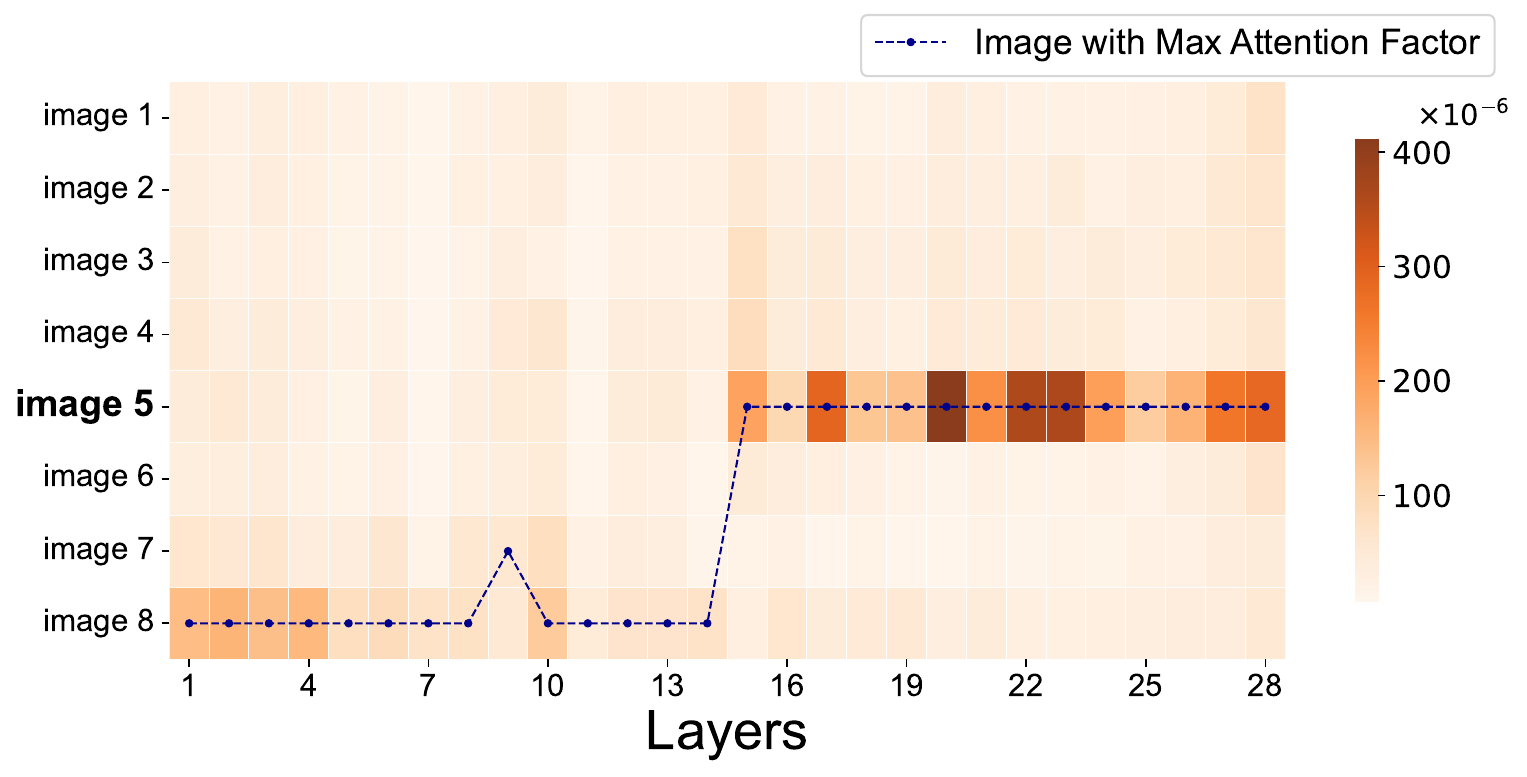}
        \label{fig:3_1_2}
    \end{subfigure}
    \tikz[baseline] \draw[line width=0.4pt, draw=black, dashed] (0,0) -- (0.95\linewidth, 0);
    \begin{subfigure}{0.45\linewidth}
        \centering
        \includegraphics[width=\linewidth]{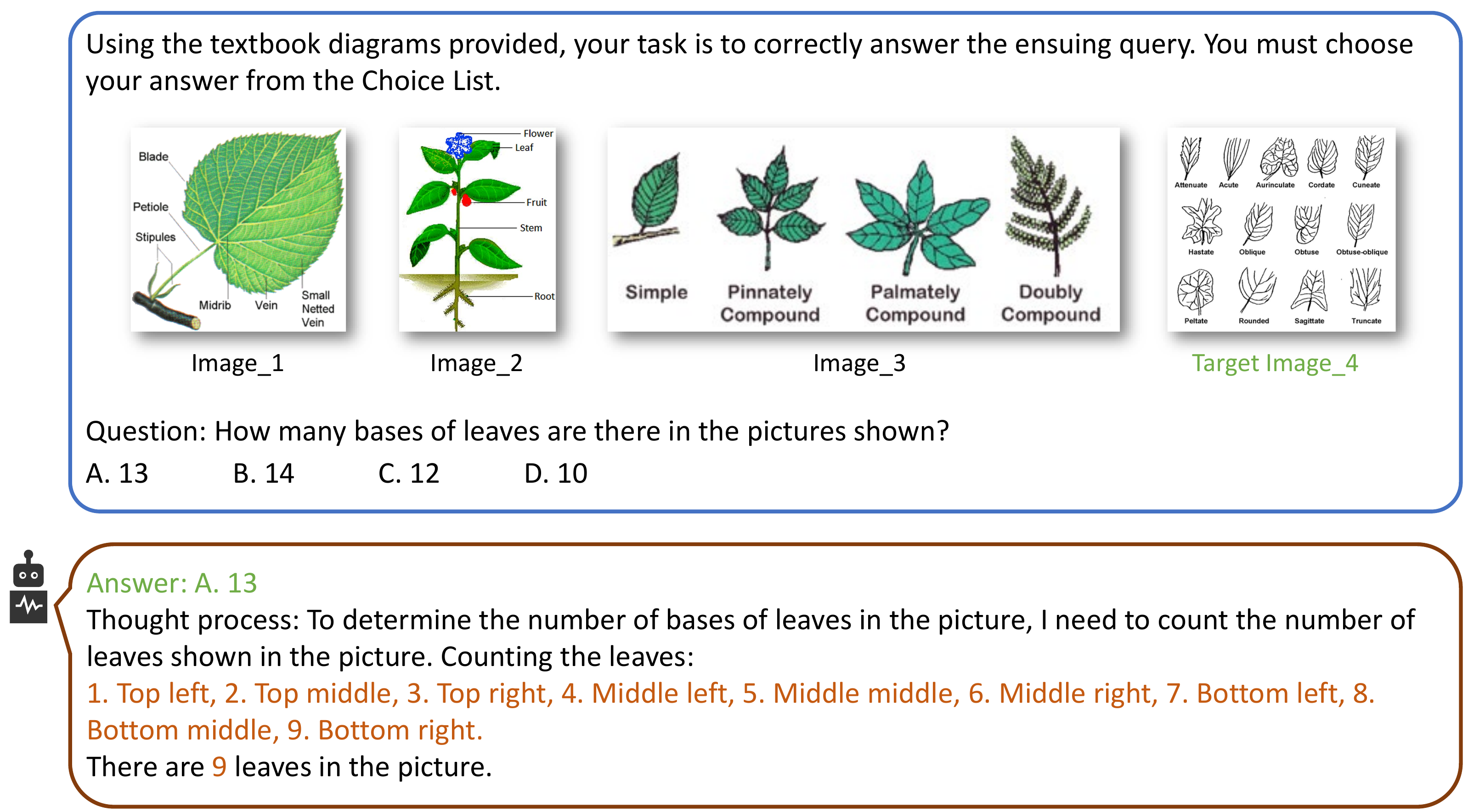}
        \label{fig:3_2_1}
    \end{subfigure}
    \hfill
    \begin{subfigure}{0.52\linewidth}
        \centering
        \includegraphics[width=\linewidth]{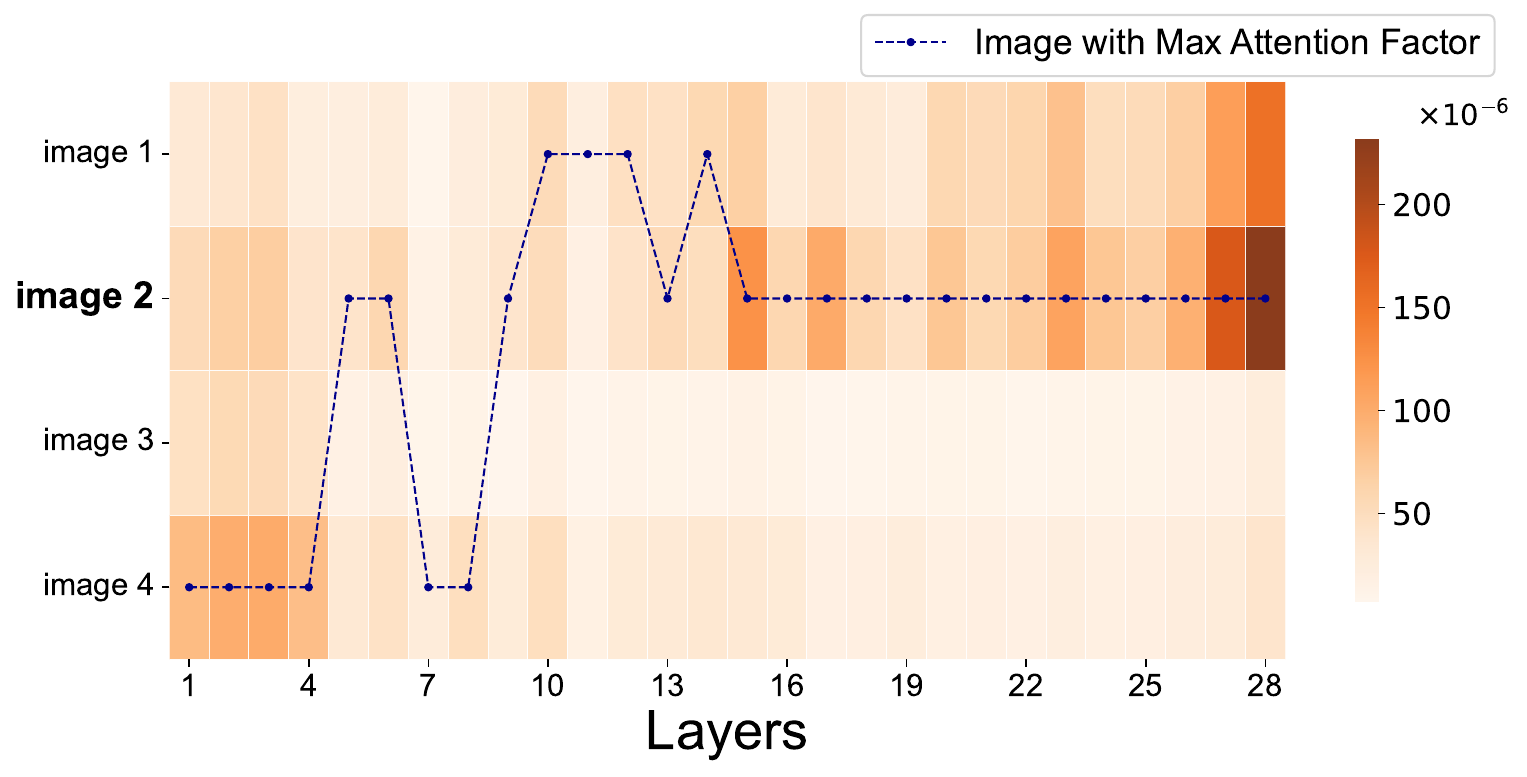}
        \label{fig:3_2_2}
    \end{subfigure}
    \vspace{-15px}
    \caption{
    \textbf{Top}: On the left, MLLM answers a caption matching question with the correct answer and explanation. On the right, the model’s attention converges on the target image.
    \textbf{Bottom}: MLLM answers an object counting question, where the fourth image corresponds to the correct answer. Despite providing the correct answer, the model’s reasoning is incorrect, showing IVMs. The heatmap reveals that the model's attention converges on a wrong image.
    }
    \vspace{-5pt}
    \label{fig:3_attn_factor_heatmap}
\end{figure*}
\section{Attention Accuracy and STME Benchmark}
In this section, we begin by an exploration of the causal attention matrices in Qwen2VL-7B \cite{qwen2vl}, the leading model within its parameter scale.
Some intriguing phenomenon observed during this analysis motivates the creation of the STME benchmark, which serves to further investigate the IVMs in MLLMs.
Using STME, we introduce attention accuracy to evaluate the degree of visual misunderstanding in MLLMs.

\subsection{Attention Distribution Phenomena}
\label{subsec:tendency}
We examine a caption-matching \cite{Flickr30k} sample with a multi-image format, as illustrated in \cref{fig:2_attn} (a).
After processing, the token sequence is systematically organized into four parts in the following order: system prompt, instruction, image, and caption (or question).
When managing interleaved image-text tokens, whether within the visual encoder or LLM's layers, Qwen2VL maintains the relative positions of tokens in the sequence \cite{qwen2vl}.
This consistency facilitates efficient extraction of tokens corresponding to text or visual inputs from the causal attention matrices.
Upon the completion of output generation, an analysis of the final attention matrices yields further insights.
For more details, please refer to the \cref{x_sec:input_format}.

\textbf{Attention matrix partition.}
For any given layer within the downstream LLM, we follow the approach of \cite{LVLM-Intrepret,vig2019analyzing} to partition the \textit{Query} ($\boldsymbol{Q}$) and \textit{Key} ($\boldsymbol{K}$) matrices into row-wise blocks.
Specifically, let $q_c$, $q_o$ denote the submatrices of the $\boldsymbol{Q}$ corresponding to input caption and the model output, respectively.
Similarly, the matrix $\boldsymbol{k_I} = \begin{bmatrix}\kappa_1,\kappa_2,\cdots,\kappa_n\end{bmatrix}^T$ represents the submatrix of $\boldsymbol{K}$ associated with $n$ input images.
As shown in \cref{fig:2_attn} (b), $q_{c}\boldsymbol{k_I}^T$ and $q_{o}\boldsymbol{k_I}^T$ correspond to the shaded regions in the attention matrix.
After applying the softmax transformation, we have:
\begin{equation}
\label{eq:softmax}
\begin{split}
&q_{c}\boldsymbol{k}_I^T
\Rightarrow
\operatorname{Softmax} \left(Attention\right)
\Rightarrow
\tilde{q}_c\tilde{\boldsymbol{k}}_I^T,\\
&q_{o}\boldsymbol{k}_I^T
\Rightarrow
\operatorname{Softmax} \left(Attention\right)
\Rightarrow
\tilde{q}_o\tilde{\boldsymbol{k}}_I^T.
\end{split}
\end{equation}
Let $\mathcal{H}$ be the index set of all attention heads, and define the concatenated query vector as $\tilde{q}_t = \left[\tilde{q}_c ; \tilde{q}_o \right]$. The attention score for the $i$-th image and the $h$-th attention head is denoted by $(\tilde{q}_t \tilde{\kappa}_i)^h \in \mathbb{R}^{m_i \times n_i}$, where $h \in \mathcal{H}$. By averaging over all heads, we define the image-attention factor $\sigma_i$ as:
\vspace{-5px}
\begin{equation}
\label{eq:attention_factor}
    \sigma_i = 
        \frac{1}{|\mathcal{H}|}
        \sum_{h\in\mathcal{H}}
        {\frac{1}{m_i n_i}}
        \sum_{j=1}^{m_i}
        \sum_{k=1}^{n_i}{(\tilde{q}_t\tilde{\kappa}_i)^{h}_{j,k}}.
\end{equation}
It is a straightforward definition to quantify the model's attention score preferences for $i$-th image in any layer.
We selected several samples for inference and computed the $\sigma_i$ value for each image across all layers.
As illustrated in \cref{fig:3_attn_factor_heatmap}, two phenomena are observed: (1) in the earlier layers, Qwen2VL demonstrates a relatively uniform attention distribution across all images; (2) in the deeper layers, the model tends to \textbf{focus  its attention on the target image}.
We hold the opinion that the first phenomenon reflects the model's initial interpretation of each image, while the shift in attention to the target image occurs once the model identifies the image pertinent to the correct answer (More results can be found in \cref{x_subsec:more_tendency}).
The comparison between the upper and lower groups in \cref{fig:3_attn_factor_heatmap} further supports the conclusion that MLLM's attention converges onto the target image if and only if there are no IVMs during inference.
Naturally, we have:
\begin{definition}
\label{def:layer-focused}
(Layer-focused image) For any given MLLM and any layer, the layer-focused image is the image with the maximum image-attention factor $\sigma$ value within that layer.
\end{definition}
Noted that the layer-focused image is defined at the layer level.
It does not imply that the whole model consistently focuses on this image.
As shown in the \cref{fig:2_attn} (c), by determining whether the layer-focused image is identical to the target image, we are able to evaluate the model's local visual misunderstandings.
However, in existing benchmarks, no dataset directly provides the association between the correct answer and the corresponding image.
To further validate the effectiveness of this idea, we design a dedicated dataset for this scenario.

\subsection{Designing Benchmark}
\label{subsec:dataset}
The dataset primarily consists of multiple-image choice questions with varying difficulty levels, and the correct answer in each sample is associated with \textbf{only one target image}.
We select eight visual tasks, each involving 2 to 20 images, mainly covering general visual contexts.
Based on MLLMs' varying performance across these tasks, we classified them into two groups: easy and hard.

\textbf{Easy group} consists of two types of tasks, with a total of 537 samples.
\begin{itemize}[noitemsep,leftmargin=*]
\vspace{-8px}
    \item Caption matching. In this task, multiple candidate images and a target image with its caption are provided, and MLLMs must identify the candidate image that matches the caption.
    Candidate images are sourced from OBELICS \cite{OBELICS}, while the target images and their captions are obtained from Flickr30k \cite{Flickr30k}.
    \item Image Needle in a Haystack \cite{Needle}.
    This task valuates the retrieval abilities of MLLMs by embedding textual data within the target image.
    The dataset for this task is taken from MileBench \cite{MileBench}.
\end{itemize}

\textbf{Hard group} consists of six multi-image tasks:
Character Order \cite{CharOrder}, Document VQA \cite{DocVQA}, Image Similarity Matching \cite{GPR1200}, Text-Rich Images QA \cite{SlideVQA}, Textbook QA \cite{TQA}, and Space Understanding \cite{nuscenes}.
Examples are provided in \cref{x_subsec:more_tendency}.
These tasks are derived from our collected data and MileBench.
However, the correct answer of the sample in the original dataset may not be tied to a single image, prompting us to develop a data filtering pipeline, through which we obtained 528 high-quality samples.

\textbf{Filtering pipeline of hard tasks.}
Initially, we remove invalid samples containing questions that can be correctly answered without relying on visual information.
Following this, we instruct GPT-4o \cite{gpt-4o} to answer these questions and identify the images related to the final answer.
Correctly answered questions are then collected, and samples with answers linked to multiple images are excluded.
Finally, a thorough manual review is conducted to ensure that all remaining samples meet the required criteria.
\begin{figure}[t]
    \centering
    \includegraphics[width=\linewidth]{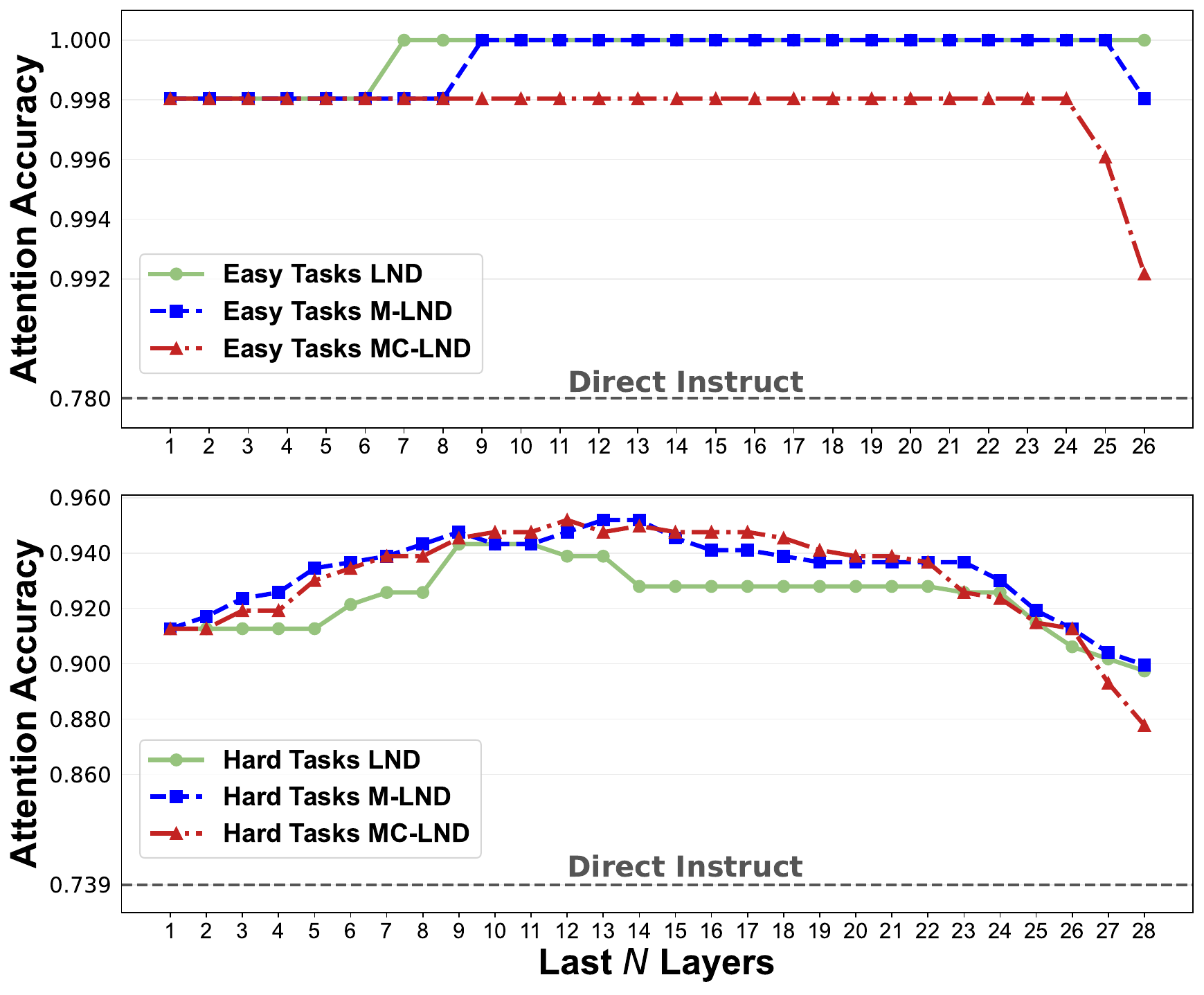}
    \vspace{-15pt}
    \caption{
     The M-LND metric demonstrates the best performance, with attention accuracy exceeding 95\% on hard tasks and achieving an astonishing 100\% on easy tasks.
     The accuracy obtained with all three metrics is significantly higher than the results from direct instructions.
     }
     \vspace{-5pt}
    \label{fig:4_metric}
\end{figure}

\subsection{From Attention to Understanding}
\label{subsec:approach}
Having constructed the dataset, we consider how to evaluate the model's IVMs.
The overall process is illustrated in \cref{fig:2_attn} (a) and (b).
Let $\tau$ denote the index of the target image, $\mathcal{I}$ the index set of all images, and $\sigma_{i,l}$ the image-attention factor for the $i$-th image within the $l$-th layer.
We utilize the image-attention factor $\sigma$ and \cref{def:layer-focused}, establishing three metrics to determine which image the model focuses on most:
\begin{itemize}[noitemsep,leftmargin=*]
\vspace{-8px}
    \item Layer-focused image of the last $N$ layers (\textbf{LND}):
    \begin{equation}
    \label{eq:LND}
        \tau = \mathop{\arg\max}\limits_{i\in\mathcal{I}} \sigma_{i,l},~~~~~l\in\mathcal{N}
    \end{equation}
    where the $\mathcal{N}$ is the index set of last $N$ layers.
    \item Mean layer-focused image of the last $N$ layers (\textbf{M-LND}):
    \vspace{-5pt}
    \begin{equation}
    \label{eq:M-LND}
        \tau = \mathop{\arg\max}\limits_{i\in\mathcal{I}} \frac{1}{N}\sum_{l\in\mathcal{N}}\sigma_{i,l}
    \end{equation}
    \vspace{-5pt}
    \item Maximum count layer-focused image of the last $N$ layers (\textbf{MC-LND}):
    \vspace{-5pt}
    \begin{equation}
    \label{eq:MC-LND}
        \tau = \mathop{\arg\max}\limits_{i\in\mathcal{I}} |\{\sigma_{i,l}: \sigma_{i,l}=\max_{k\in\mathcal{I}}\sigma_{k,l},~l\in\mathcal{N}\}|
    \end{equation}
\end{itemize}
\vspace{-5pt}
Building on these three metrics, we can determine which image the model concentrates on.
\begin{definition}
\label{def:model-focused}
(Model-focused Image) For any given MLLM, the model-focused image is the one corresponding to the maximum value among the LND, M-LND, and MC-LND metrics.
\end{definition}
The maximum value is used as the final evaluation criterion to reflect the upper bound of each model.
In practice, different metrics can be applied, and as shown in \cref{fig:4_metric} and \cref{x_sec:other_model_metric}, the differences are marginal.
For inference on a single sample, \cref{def:model-focused} operates at the model level. By comparing the target image with the model-focused image, we can determine whether IVMs occur during the inference process.
\begin{definition}
\label{def:attn_correct}
(Attention Correctness) For any given MLLM and sample with single target image, the model's attention is correct if the model-focused image is identical to the target image.
\end{definition}
Subsequently, we evaluate Qwen2VL-7B on the STME, using Chain-of-Thought \cite{CoT} prompts to guide the model.
Correctly answered samples are selected to calculate the attention accuracy based on \cref{def:attn_correct}.
To validate the effectiveness of our method, we also directly instruct model to output the index of the target image.

\textbf{Results and analysis.}
As illustrated in \cref{fig:4_metric}, the Qwen2VL-7B achieves a remarkable 100\% attention accuracy on easy tasks (The results for other models are presented in \cref{x_sec:other_model_metric}).
Compared to directly instructing the model to output the index of the target image, the accuracy of our metrics is significantly higher.
In other words, for every correctly answered sample, it consistently focuses on the target image, indicating no IVMs.
To further validate the experimental results, we utilize GPT-4o \cite{gpt-4o} to evaluate the correct CoT responses of Qwen2VL-7B.
The results show that its reasoning process of each sample is also correct, providing strong validation for the effectiveness of attention accuracy.
More comprehensive and thorough experiments will be conducted in the following section.
\section{Experiments}
\label{sec:experiments}
\begin{figure*}[t]
    \centering
    \includegraphics[width=\linewidth]{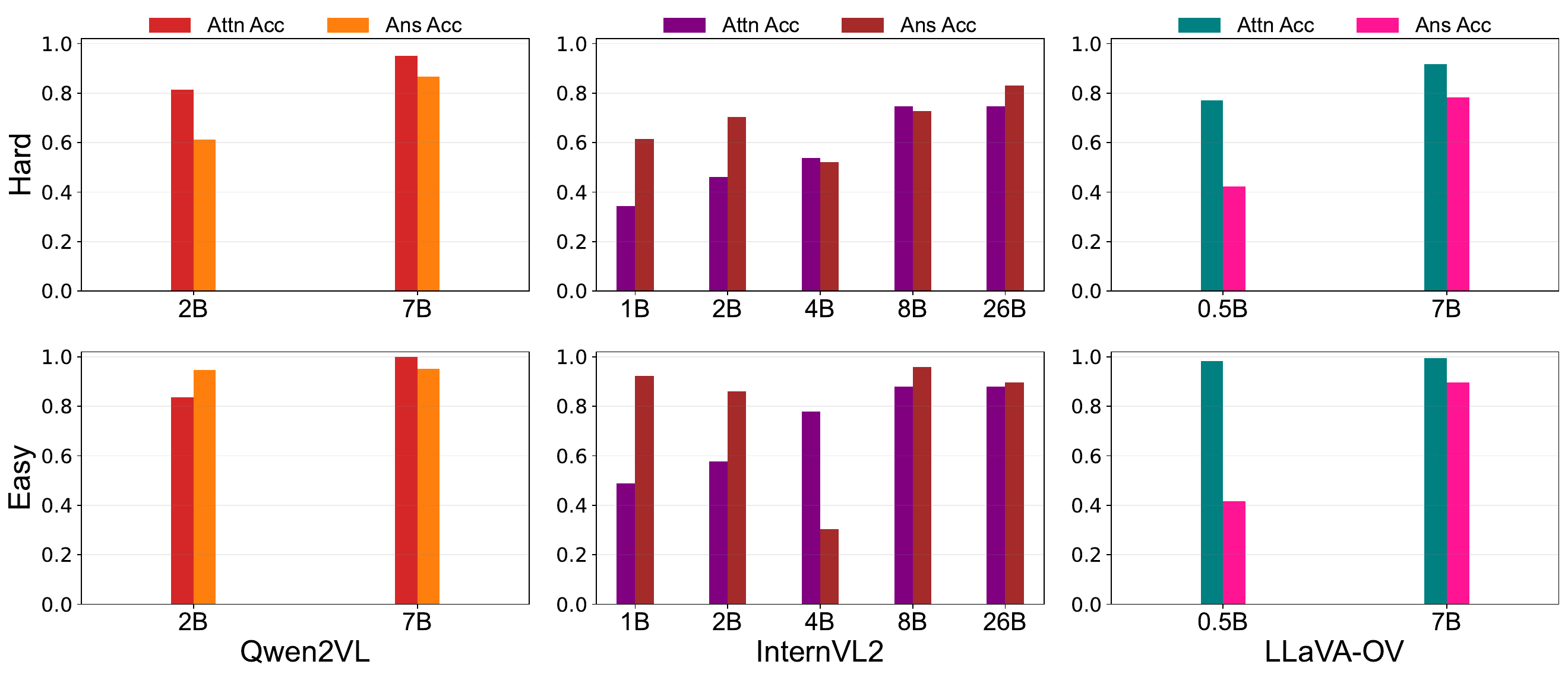}
    \vspace{-20px}
    \caption{
    As the model scale increases, the attention accuracy also improves, with models of varying scales exhibiting particularly high attention accuracy on less challenging tasks.
    In contrast, answer accuracy does not follow the same trend.
    This indicates an enhancement in the model's visual capabilities, but due to constraints in the downstream LLM, no corresponding performance improvement is observed.
    }
    \vspace{-3px}
    \label{fig:5_diverse_scales}
\end{figure*}
In this section, the proposed approach is applied to models from different series and scales, with inference tasks of varying difficulty.
This is followed by an in-depth analysis of positional biases.
Finally, the approach is expanded to the token level.
Due to space limitations, more results including studies on hallucinations and experiments that provide indirect validation of the method's effectiveness, are presented in \cref{x_sec:supp_exp}.

\subsection{Experiments Setup}
We consider Qwen2VL \cite{qwen2vl}, InternVL2 \cite{internvl2}, and LLaVA-OneVision \cite{llava-ov}.
All three series models are capable of understanding multiple images and interleaved image-text information.
The inference mode remains consistent with the the methodology outlined in \cref{subsec:tendency}.
We use the CoT paradigm to guide the model's responses on both easy and hard tasks.

\textbf{Details.}
Qwen2VL series and InternVL2 series models utilize dynamic resolution, mapping different images to varying numbers of tokens.
Due to the limited GPU memory, the maximum number of tokens varies depending on the number of total images included in the sample.
LLaVA-OneVision series models resize all images to a fixed size, which means that each row and column vector in the attention matrix corresponds directly to the patches of the original images.
Consequently, we can further extend our approach to a more granular level.
For more experimental details, please refer to \cref{x_subsec:exp_detail}.

\textbf{Evaluation.}
We evaluate the models from two perspectives: answer accuracy and attention accuracy.
The former demonstrates the models' capability for visual understanding, and the later reflects the degree of IVMs.
Similar to \cref{subsec:approach}, attention accuracy is calculated on correctly answered samples using the LND, M-LND and MC-LND metrics across different $N$.
The highest attention accuracy obtained is used as the final value.
All inference processes are conducted on H100 GPUs.
\begin{table}
  \centering
  \resizebox{\linewidth}{!}{
  \begin{tabular}{@{}lcccccc@{}}
    \toprule
    \multirow{2}{*}{} & \multirow{2}{*}{Params} & \multicolumn{2}{c}{\textbf{Attn Acc (\%)}} & \multicolumn{2}{c}{\textbf{Ans Acc (\%)}} \\
    \cmidrule(lr){3-4} \cmidrule(lr){5-6}
    & & Easy & Hard & Easy & Hard \\
    \midrule
    Qwen2VL \cite{qwen2vl} & 7B &\textbf{100} & \textbf{95.2} & \textbf{95.2} & \textbf{86.7} \\
    InternVL2 \cite{internvl2} & 8B & 96.0 & 72.8 & 87.9 & 74.6 \\
    LLaVA-OV \cite{llava-ov} & 7B  & 99.6 & 91.8 & 89.6 & 78.2 \\
    \bottomrule
  \end{tabular}
  }
  \vspace{-3px}
  \caption{The difference in attention accuracy indicates notable disparities in visual capabilities across all models.
  However, powerful downstream LLMs provide some correction, making the final answer accuracy appear less divergent.}
  \vspace{-8px}
  \label{tab:diverse_series}
\end{table}

\subsection{IVM Analysis Across Different Model Series}
\label{subsec:diverse_series}
We compare Qwen2VL, InternVL2, and LLaVA-OV.
Considering the broad applicability and overall performance, we choose models with 7B to 8B parameters for our study.
Close parameter scale provide a fair comparison of IVM levels across different model series.

\textbf{Main results.}
As listed in \cref{tab:diverse_series}, Qwen2VL-7B achieves the highest attention accuracy, indicating its lowest degree of IVMs.
Although LLaVA-OV and InternVL achieve comparable answer accuracy, the more advanced LLaVA-OV demonstrates higher attention accuracy.
This highlights a notable difference in their levels of IVMs and suggests that the visual capabilities of LLaVA-OV are significantly stronger than those of InternVL.
According to the scores of these three models on currently available benchmarks \cite{qwen2vl, internvl2, llava-ov}, we attribute this to differences in training sufficiency and balance.
Therefore, we believe the attention accuracy can serve as an internal guide for optimizing the training of MLLMs.

\textbf{Impact of task difficulty.}
Attention accuracy varies more for hard tasks than easy tasks across all models.
Challenging tasks have a lower tolerance for IVMs, amplifying performance differences across models.
Easy tasks can be handled with greater ease, resulting in less pronounced differences.
However, an anomaly appears in \cref{tab:diverse_series}:
answer accuracy of InternVL on easy tasks is only 87.9\%, notably lower than the Qwen2VL.
Therefore, we analysis its reasoning process and find this issue may stem from limitations in its visual encoder or data preprocessing.
Specifically, the easy tasks include the ``Needle In A Multimodal Haystack'' \cite{Needle} task in which InternVL easily locate the target image, thereby achieving high attention accuracy.
On the other hand, its lower answer accuracy may result from improperly segmented image patches during preprocessing or limited OCR capabilities in the visual encoder, which prevents accurate recognition of all numerical information.
This observation suggests that combining answer accuracy with attention accuracy offers a more comprehensive assessment of MLLMs.

\begin{figure*}[t]
    \centering
    \includegraphics[width=\linewidth]{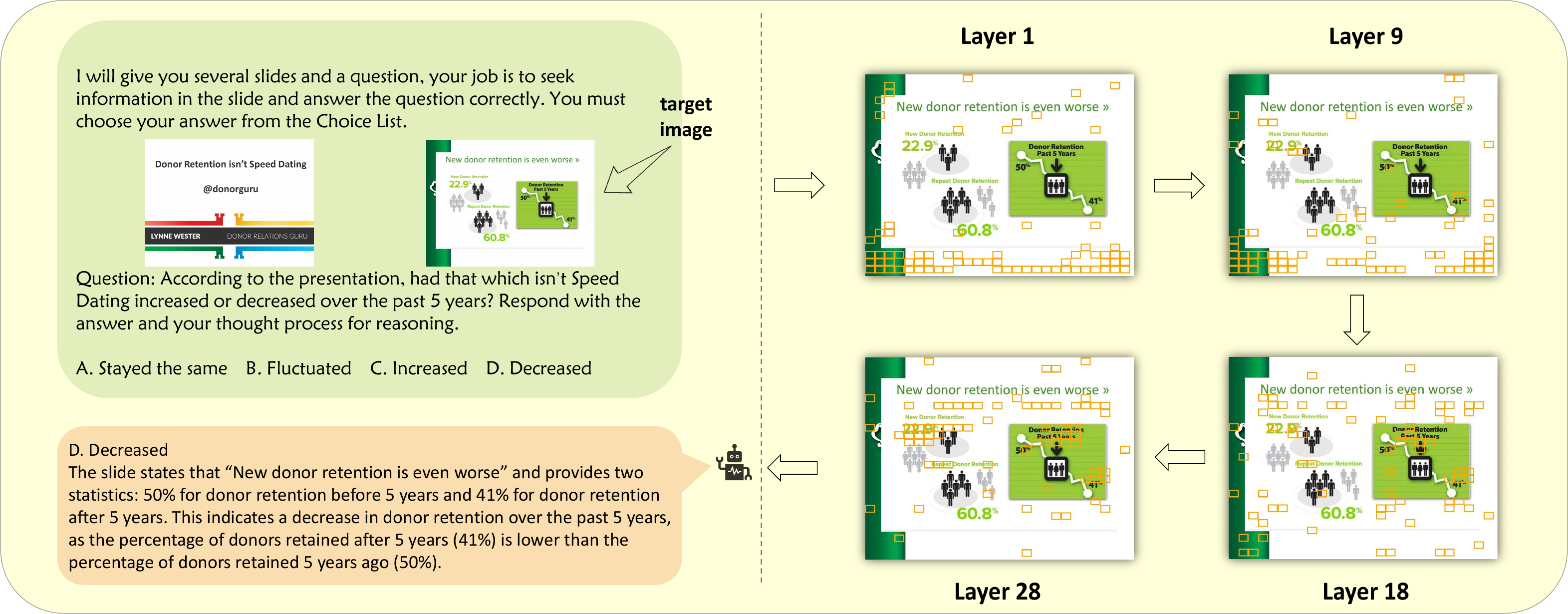}
    \caption{\textbf{Left}: Using CoT prompting, LLaVA-OneVision-7B is guided to reason through the SlideVQA task.
    The model successfully answers the question and focuses on the target image.
    \textbf{Right}: The patch-attention factor values for the target image are computed across all layers. We highlight the patches with the top 10\% patch-attention values in orange boxes.
    As the layer deepens, the model progressively focuses on regions of the image containing information such as ``50\%'' and ``41\%'', which are directly related to the correct answer.}
    \label{fig:6_token_attn}
\end{figure*}

\subsection{The Effect of Model Scale}
We evaluate the models of different sizes within the three series, with results shown in \cref{fig:5_diverse_scales}.
As the model parameter scale increases, attention accuracy consistently improves, suggesting stronger visual capabilities and lower degree of visual misunderstandings.
In contrast, answer accuracy does not follow this trend.

\textbf{Comparative case analysis.}
In the second column of subplots in \cref{fig:5_diverse_scales}, the answer accuracy of InternVL2 models does not positively correlate with model scale.
Closer analysis reveals that this inconsistency arises from instruction-following failures and disorganized responses, likely sourced from limitations within the downstream LLMs.
Since the InternVL2 models of varying scales incorporate different downstream LLMs, we attribute the observed differences in answer accuracy to unaligned knowledge embeddings.
This suggests that, under similar training data and methodologies, the degree of IVMs indeed decreases as model scale grows, even when the model architectures differ.

\subsection{Positional Bias Independence}
\begin{table}
  \centering
  \resizebox{\linewidth}{!}{
  \begin{tabular}{@{}lcccccc@{}}
    \toprule
    \multirow{2}{*}{} & \multirow{2}{*}{Params} & \multicolumn{2}{c}{\textbf{Attn Acc (\%)}} & \multicolumn{2}{c}{\textbf{Ans Acc (\%)}} \\
    \cmidrule(lr){3-4} \cmidrule(lr){5-6}
    & & Easy & Hard & Easy & Hard \\
    \midrule  
    Qwen2VL \cite{qwen2vl} & 2B & 85.0 ($\pm$1.3) & 82.3 ($\pm$0.8) & 91.1 ($\pm$4.5) & 60.1 ($\pm$4.4) \\
    Qwen2VL \cite{qwen2vl} & 7B & 100 ($\pm$0.0) & 95.2 ($\pm$0.6) & 92.8 ($\pm$3.3) & 83.8 ($\pm$5.2) \\
    InternVL2 \cite{internvl2} & 2B & 88.6 ($\pm$0.9) & 68.9 ($\pm$1.6) & 59.0 ($\pm$4.3) & 48.3 ($\pm$2.9) \\
    InternVL2 \cite{internvl2} & 8B & 95.7 ($\pm$0.3) & 72.4 ($\pm$0.3) & 87.2 ($\pm$4.3) & 72.6 ($\pm$4.0) \\
    LLaVA-OV \cite{llava-ov} & 0.5B  & 97.2 ($\pm$1.0) & 77.6 ($\pm$0.5) & 44.8 ($\pm$6.1) & 42.2 ($\pm$8.2) \\
    LLaVA-OV \cite{llava-ov} & 7B  & 99.1 ($\pm$0.5) & 91.5 ($\pm$0.6) & 85.9 ($\pm$4.4) & 80.2 ($\pm$3.1) \\
    \hline
    Avg Variance & & $\pm$\textbf{0.67} & $\pm$\textbf{0.73} & $\pm$4.45 & $\pm$4.63 \\
    \bottomrule
  \end{tabular}
  }
  \vspace{-3px}
  \caption{Shuffling the image order to eliminate positional bias. The variance in attention accuracy is smaller, while the variance in answer accuracy is much greater. This indicates that attention accuracy is more stable and unaffected by positional bias.}
  \vspace{-8px}
  \label{tab:positional_bias}
\end{table}
To examine the impact of image order on attention accuracy, we randomly shuffle the image sequences within both easy and hard tasks.
The target image's position is altered compared to its original placement.

After shuffling the image order five times, we conducted inferences across different models, yielding the results shown in \cref{tab:positional_bias}.
It is evident that the attention accuracy metric is minimally affected by positional bias.
Across all models, the average variance of attention accuracy does not exceed 1\%, demonstrating the \textbf{robustness} of attention accuracy.
The instability of answer accuracy limits the comparison between different models.
In this case, attention accuracy serves as an excellent complementary metric.
\begin{figure*}[t]
    \centering
    \includegraphics[width=\linewidth]{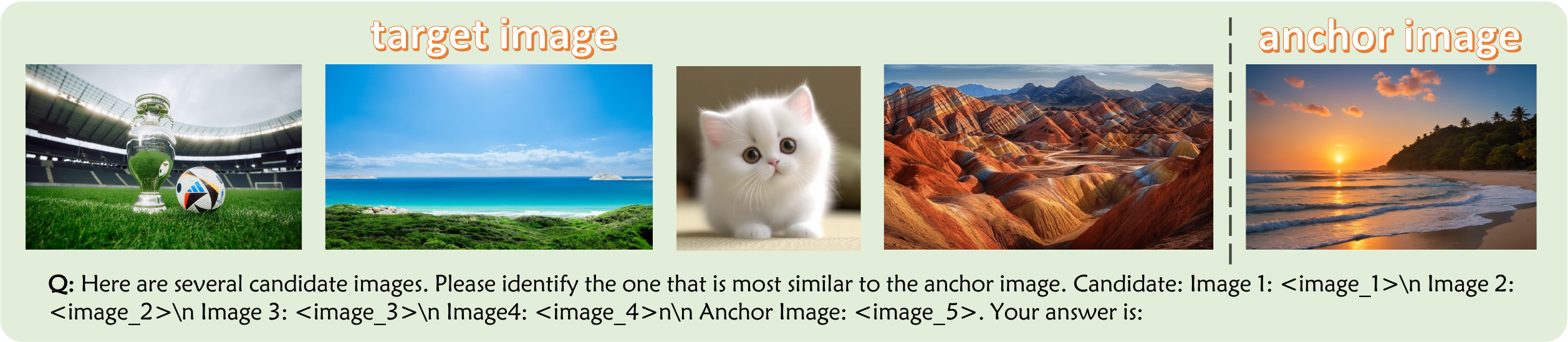}
    \vspace{-15px}
    \caption{
    Image-to-image similarity matching task.
    We place the anchor image last, enabling the calculation of the attention factor.
    The second image is correct answer.
    }
    \vspace{-5px}
    \label{fig:7_img_img}
\end{figure*}
\section{Approach Extensions}
In this section, we delve into the attention trends mentioned in \cref{subsec:tendency} at patch level, and build the intrinsic relationships between the vectors corresponding to image patches, further generalizing the method presented in \cref{subsec:approach}.

\subsection{Patch-level Multimodal Connections}
We choose the LLaVA-OneVision-7B \cite{llava-ov} for this analysis.
Unlike models with dynamic resolution, it resizes all images to a uniform size, which allows us to associate the row and column vectors in the attention matrix with patches of the original images.

\textbf{Definition of Patch-attention Factor.}
Let $\nu_{i,n}$ denote the row vector of the \textit{key} matrix in \textit{Attention} module \cite{transformer} corresponding to the $n$-th patch of the $i$-th image.
Additionally, let $\mathcal{H}$ be the index set of all heads, and $(\tilde{q}_t \nu_{i,n})^h \in \mathbb{R}^{m_i \times 1}$ represents the attention score for the $i$-th image in the $h$-th attention head.
Similar to \cref{eq:attention_factor}, we define the patch-attention factor as follows:
\begin{equation}
\label{eq:patch_attention_factor}
    \rho_{i,n} = 
        \frac{1}{|\mathcal{H}|}
        \sum_{h\in\mathcal{H}}
        {\frac{1}{m_i}}
        \sum_{j=1}^{m_i}{(\tilde{q}_t \nu_{i,n})^{h}_{j}}.
\end{equation}
In a similar manner, we hold the opinion that $\rho$ can be applied to determine whether the MLLM is focused on a specific patch within an image.
This extension allows for a more granular evaluation of the implicit visual errors in MLLMs, especially in complex visual scenes and tasks that require careful attention to image details.

To validate our hypothesis, we tasked LLaVA-OV with a image-text interleaved reasoning task.
As illustrated in the left portion of \cref{fig:6_token_attn}, the model correctly answers and effectively focuses on the target image.
We then calculated the patch-attention factor for each patch in the target image across all layers of the downstream LLM.
As shown in the right portion of \cref{fig:6_token_attn}, we observe a progressive increase in focus on the useful information from shallow to deep layers, with attention being continuously redistributed.
This phenomenon was consistently observed across a variety of tasks in our experiments.

\begin{table}
  \centering
  \resizebox{0.8\linewidth}{!}{
  \begin{tabular}{@{}lcccc@{}}
    \toprule
    \multirow{2}{*}{} & \multirow{2}{*}{Params} & \multicolumn{2}{c}{\textbf{Attn Acc (\%)}} \\
    \cmidrule(lr){3-4}
    & & img-img & txt-img \\
    \midrule
    Qwen2VL \cite{qwen2vl} & 2B & \textbf{95.1} & 92.3 \\
    Qwen2VL \cite{qwen2vl} & 7B & \textbf{99.6} & 97.7 \\
    InternVL2 \cite{internvl2} & 2B & \textbf{56.3} & 56.2 \\
    InternVL2 \cite{internvl2} & 8B & \textbf{57.2} & 56.9 \\
    LLaVA-OV \cite{llava-ov} & 0.5B  & \textbf{80.9} & 77.7 \\
    LLaVA-OV \cite{llava-ov} & 7B  & \textbf{85.8} & 83.9 \\
    \bottomrule
  \end{tabular}
  }
  \caption{In the image-to-image similarity matching task, using unimodal interleaved regions generally leads to higher attention accuracy.}
  \vspace{-5px}
  \label{tab:image-image}
\end{table}
\subsection{Interwoven Visuals: Attention as the Link}
\label{subsec:image_token}
In the previous sections, we focused on the dependencies between different modalities.
Here, we examine how images interact with each other within the attention matrix.

\textbf{Dataset preparation.}
As shown in \cref{fig:7_img_img}, we first consider an image-to-image similarity matching task: the MLLM is provided with several candidate images and one anchor image, and is instructed to select the candidate image most similar to the anchor image.
The images are sourced from the OBELICS \cite{OBELICS} and GPR1200 \cite{GPR1200}, covering a wide range of categories such as daily scenes, art, diagrams, flora and fauna, totaling 270 samples.
The target image are carefully selected to share obvious features with the anchor image, making it easy for a human to identify the correct answer at a glance.

\textbf{Extrcting submatrix.}
Distinguish from \cref{subsec:tendency}, we consider the interaction between the anchor image and each candidate image.
The extracted submatrix is as follows:
\begin{equation}
\label{eq:attn_sub_image_to_image}
    \boldsymbol{Attn}_{sub}
    = \tilde{q}_a * \begin{bmatrix}\tilde{\kappa}_1,\tilde{\kappa}_2,\cdots,\tilde{\kappa}_{n-1}\end{bmatrix},
\end{equation}
where $\tilde{q}_a$ is the \textit{Query} submatrix corresponding to the anchor image, and $\tilde{\kappa}_{i}$ is the \textit{Key} submatrix corresponding to the candidate image.
Following the \cref{subsec:tendency} and \cref{subsec:approach}, we calculate the image-attention factor and use the LND, M-LND, and MC-LND metrics to obtain attention accuracy.

\textbf{Results of the experiments.}
The results in \cref{tab:image-image} indicate that, in the unimodal setting, attention scores also tend to concentrate on the target image, with the computed attention accuracy reaching even higher levels.
This suggests that the methods for calculating attention accuracy are diverse and applicable to a wide range of scenarios.
\section{Related Work}
The ability to process and understand multiple images is a critical aspect of MLLMs. Closed-source models \cite{gpt-4v,gemini1_5pro,claude3,glm-4v} perform strongly on multi-image benchmarks \cite{MMDU,MMIU}.
Open-source models \cite{CogVLM2,mantis,internlm-xcomposer} have also made significant progress, especially Qwen2VL \cite{qwen2vl}, which achieves impressive results on various visual tasks by using token-level dynamic resolution.

The evaluation of MLLMs' visual capabilities has garnered significant attention.
BLINK \cite{BLINK} consists of tasks that are easy for humans but challenging for models.
Some works \cite{CMM,SID,DRESS} primarily address hallucinations, while others \cite{MuirBench, MMIE} focus on the models' ability to handle long-context visual scenarios.
The broad range of world knowledge \cite{MMMU, CMMU} has also drawn attention.
Each of these studies offers a unique perspective on evaluating the visual capabilities of MLLMs.
\section{Conclusion}
We contribute the STME benchmark, which encompasses a range of visual tasks and is adaptable for evaluating visual misunderstandings in models.
To assess the attention allocated to visual information, we establish both layer-level and model-level metrics, with attention accuracy serving as a key measure of implicit visual misunderstandings.
Experiments demonstrate the effectiveness of our method across a variety of models.
Compared to traditional methods that focus solely on explicit visual misunderstandings, attention accuracy provides a more direct and reliable evaluation of a model's visual capabilities.
Finally, we extend our approach in two ways: conducting a more granular layer-level analysis and exploring relationships within the same modality.

We believe this method is highly versatile, with potential applications in LLMs and other fields.
Due to its equivariant property, attention accuracy can consistently evaluate both pretrained and fine-tuned models on a unified scale.
Future work will further explore the broader applicability of this method across various tasks and domains.

\bibliography{main}
\bibliographystyle{icml2025}

\newpage
\appendix
\onecolumn
\section{Discussion}
\subsection{Limitations}
We establish both layer-level and model-level metrics, with attention accuracy serving as a key measure of IVMs in MLLMs on our proposed STME benchmark.
While this metric evaluates models from a purely visual perspective and is robust to image positional bias, we do not explore methods for mitigating IVMs within the models themselves.
Additionally, we have not explore models at various training stages, such as pretraining, SFT, DPO, or RL training, to investigate their effects on attention accuracy.

Although we have extensively analyzed the differences in attention accuracy across models on diverse inference data, further granular analysis remains possible.
Moreover, several mechanisms within the Attention module have not been fully explored, which could offer valuable insights into the visual capabilities of MLLMs.
In terms of engineering, our approach necessitates modifications to the model's structural code, adding practical complexity to its implementation.

\subsection{Expectations}
Attention accuracy complements existing MLLM visual capability evaluation systems by distinguishing whether a model’s deficiencies originate from the downstream LLM or its visual components.
This distinction can guide training data selection and the development of methodologies to mitigate IVMs in MLLMs.
Due to its equivariant property, attention accuracy enables consistent evaluation of both pretrained and post-trained models on a unified scale.
By analyzing these differences, we can assess the impact of various training methods on models purely from a visual perspective, leading to deeper insights.

Unlike traditional evaluation methods, attention accuracy examines whether a model effectively attends to target visual information by leveraging its internal mechanisms.
This approach can be extended to other multimodal scenarios, such as text-audio or vision-audio tasks.
Moreover, using attention accuracy to filter data for more diverse training strategies presents a promising research direction.
Fundamentally, this method clusters data based on the model’s internal attention distribution.

By evaluating models' visual capabilities through their internal mechanisms for the first time, we hope our work will inspire further innovations in vision models.
Our dataset is available at \url{https://huggingface.co/datasets/bestpf/STME}, and the corresponding code can be accessed at \url{https://github.com/WellDonePF/STME}.

\newpage
\section{Image-attention factor calculation}
\label{x_sec:input_format}
\begin{figure}[htbp]
    \centering
    \includegraphics[width=0.95\linewidth]{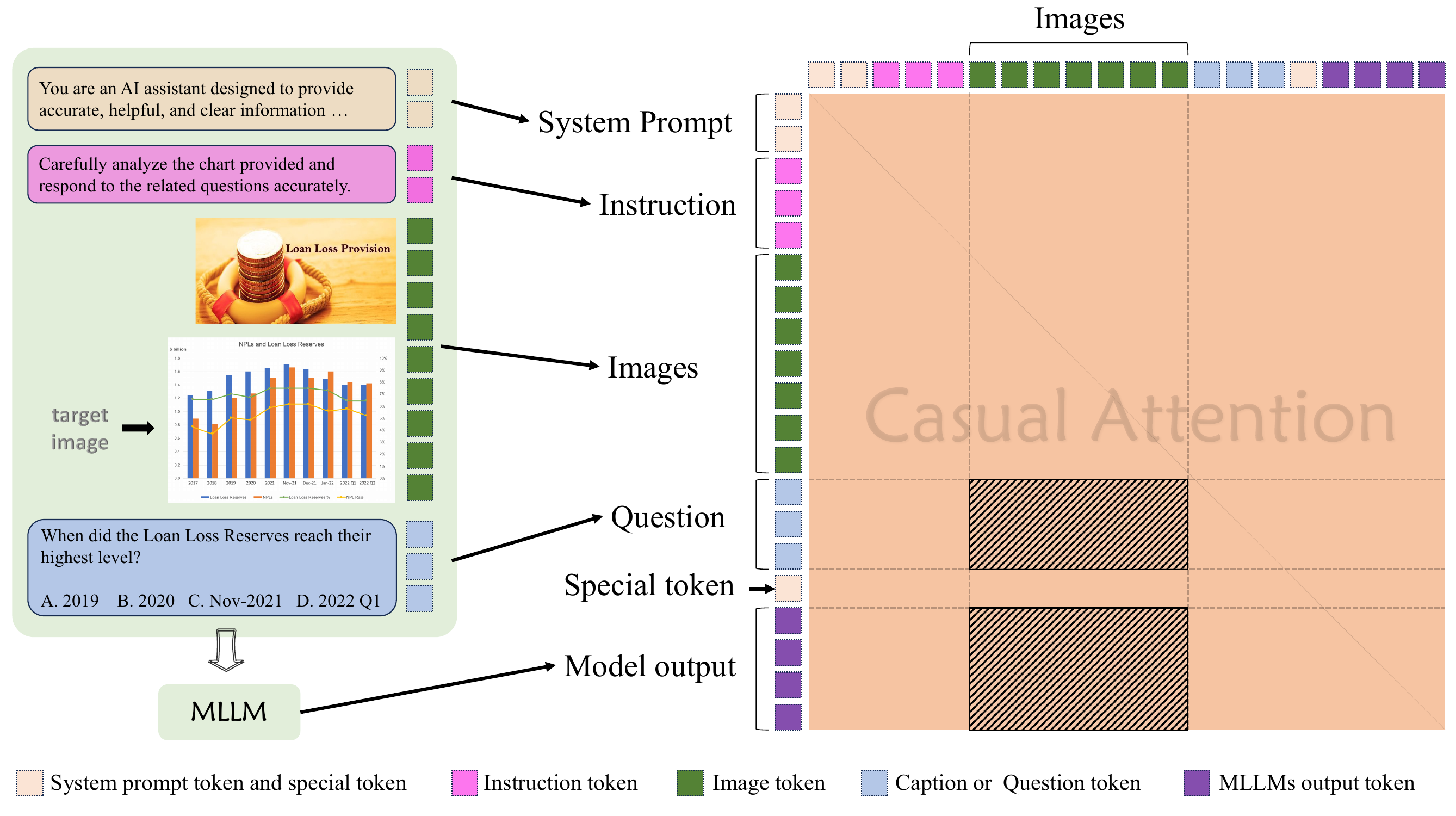}
    \caption{The order and position of input and output tokens in the causal attention matrix. The shaded submatrices are used to calculate the image-attention factor.
    }
    \label{x_fig:8}
\end{figure}

We provide a more detailed breakdown of all token types, as illustrated in \cref{x_fig:8}.
These include system prompt tokens, special tokens, instruction tokens, image tokens, target tokens, and model output tokens.
Depending on the specific task, the target tokens can be categorized into three types:
\begin{itemize}[noitemsep,leftmargin=*]
    \item Caption tokens are used in caption matching tasks.
    \item Question tokens correspond to questions and answer options in non-caption visual tasks.
    \item Anchor image tokens as described in \cref{subsec:image_token}.
\end{itemize}
When calculating the image-attention factor $\sigma$ within a layer, the vectors corresponding to the system prompt tokens, special tokens, and instruction tokens are excluded.

There are a total of eight task types, classified as either easy or hard tasks.
The token order and types for all tasks follow the structure shown in \cref{x_fig:8}, ensuring that the corresponding image-attention factor, $\sigma$, can be computed based on \cref{eq:attention_factor}.
Our computation process involves extracting two submatrices after applying the softmax transformation.
The first submatrix corresponds to the dot product between the row vectors from the \textit{Query} matrix (associated with the target token) and the column vectors from the \textit{Key} matrix (corresponding to the image token).
The second submatrix is derived from a similar operation, where the row vectors correspond to the model's output text token.

We partition matrices $\boldsymbol{Q}$ and $\boldsymbol{K}$ into blocks by rows.
\begin{equation}
    \boldsymbol{Q} = \begin{bmatrix}
        \cdots \\
        q_c \\
        \cdots \\
        q_o
    \end{bmatrix},~~~~~~~~~
    \boldsymbol{K} = \begin{bmatrix}
        \cdots \\
        \boldsymbol{k}_I \\
        \cdots
    \end{bmatrix},
\end{equation}
Next, we partition the attention matrix of all heads into blocks as follows.
\begin{equation}
\begin{split}
    \boldsymbol{Attn}
    &= \operatorname{Softmax}(\frac{\boldsymbol{Q}*\boldsymbol{K}^T}{\sqrt{d}}) \\
    &= \operatorname{Softmax} \left(
        \begin{bmatrix}
            \cdots & \cdots & \cdots \\
            \cdots & q_{c}\boldsymbol{k}_I^T & \cdots \\
            \cdots & \cdots & \cdots \\
            \cdots & q_{o}\boldsymbol{k}_I^T & \cdots
        \end{bmatrix} / \sqrt{d}
    \right).
\end{split}
\end{equation}
Here $q_{c}\boldsymbol{k}_I^T$ and $q_{o}\boldsymbol{k}_I^T$ correspond to the shaded regions in the attention matrix shown in \cref{x_fig:8} (b), and $d$ represents the embedding dimension.
After \cref{eq:softmax}, these two components are extracted and concatenated:
\begin{equation}
\label{eq:attn_sub}
\begin{split}
    \boldsymbol{Attn}_{sub}
    &= \begin{bmatrix}
        \tilde{q}_c \\
        \tilde{q}_o
    \end{bmatrix} * \tilde{\boldsymbol{k}}_I^T
    = \tilde{q}_t * \begin{bmatrix}\tilde{\kappa}_1,\tilde{\kappa}_2,\cdots,\tilde{\kappa}_n\end{bmatrix}, \\
    & \text{where}~~~ \tilde{q}_t=\begin{bmatrix}
                            \tilde{q}_c \\
                            \tilde{q}_o
                        \end{bmatrix}.
\end{split}
\end{equation}
Then the \cref{eq:attention_factor} is derived.

\newpage
\section{Supplementary Experiments}
\label{x_sec:supp_exp}
\subsection{Implementation Details}
\label{x_subsec:exp_detail}
The Qwen2VL series models utilize dynamic resolution, achieved through dynamic resizing, pixel reorganization, and a specially designed visual encoder, which maps different images to varying numbers of tokens.
We set the minimum resolution for all images after dynamically resizing to $256\times28\times28$, and the maximum resolution varies depending on the number of images in the sample, ranging from $256\times28\times28$ to $426\times28\times28$.
Considering hardware memory constraints, the resolution of each image is determined by the image number in the question.

The InternVL2 series models also employ dynamic resolution, but with a different approach.
Initially, sub-images are selected based on the aspect ratio of the images.
These sub-images, along with an optional overall thumbnail, are then resized to $448 \times 448$ and concatenated together.
All these images are treated as tokens representing the complete image, and the image-attention factor $\sigma$ values are calculated collectively.
The maximum number of sub-images varies depending on the total number of images included in the sample.
For models with a scale not exceeding 8B, the maximum number of sub-images is set to $3\sim6$; for larger models, it is set to $1\sim6$.

The LLaVA-OneVision series models resize all images to a fixed size, which means that each row and column vector in the attention matrix corresponds directly to the patches of the original images.

During inference, we adopt a greedy mode to minimize the disturbances caused by random uncertainty.
For some models using Qwen2 \cite{qwen2} as the downstream LLM, due to issues in the source code implementation, we increase the precision of the \textit{Query}, \textit{Key}, and \textit{Value} matrices to 32-bit in the Attention module.

Current LLMs use the KV cache method during inference to reduce computational load, thereby accelerating inference and reducing memory usage.
When analyzing attention scores, we first perform a full inference and, for each newly generated token, concatenate the corresponding tensor to the bottom of the original attention matrix.
A zero vector is then concatenated to the right side of the matrix to ensure it remains square.

\subsection{Evaluation of Hallucinations}
\label{x_subsec:hallucinations}
\begin{figure}[htbp]
    \centering
    \includegraphics[width=0.6\linewidth]{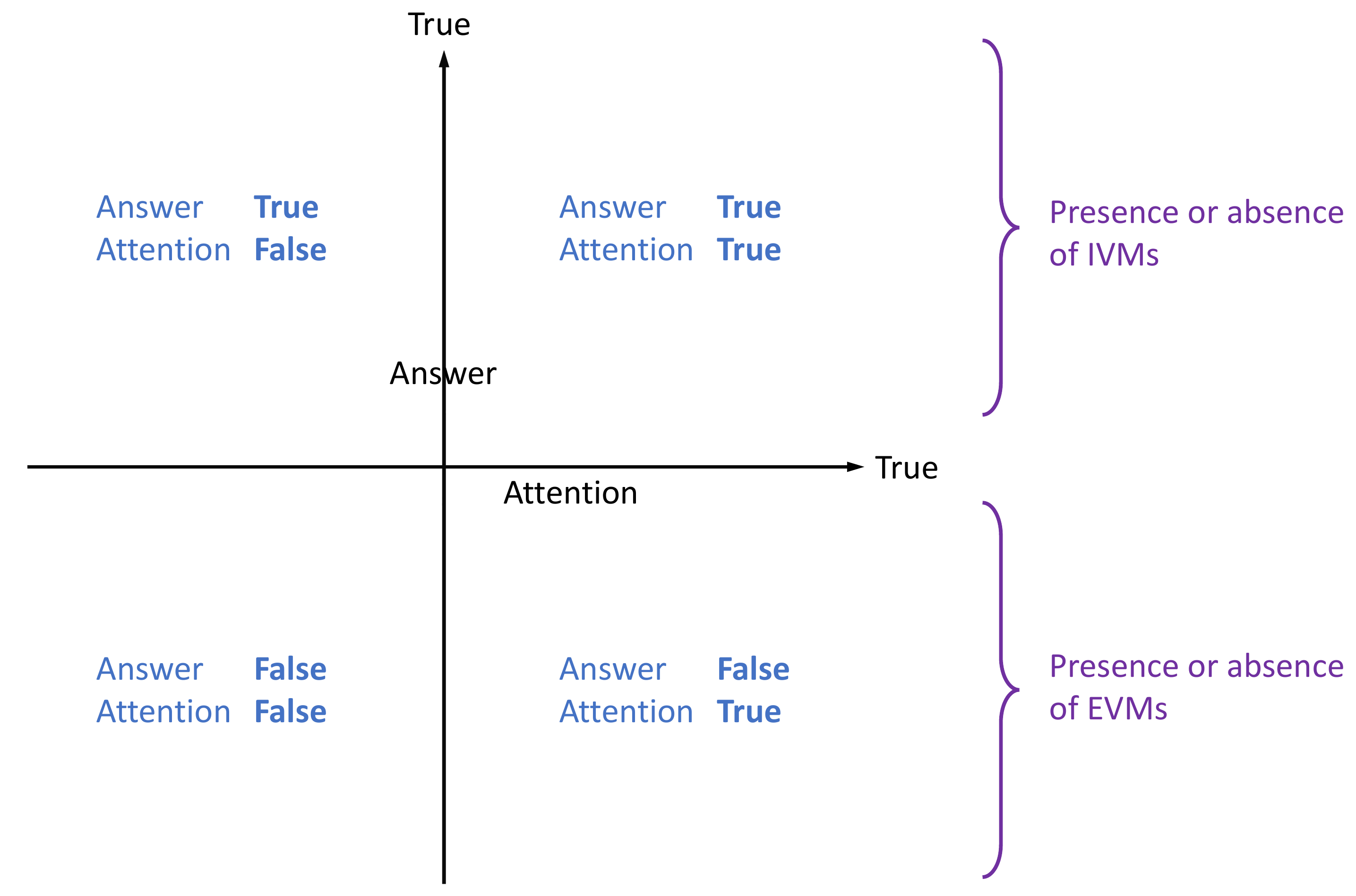}
    \caption{Attention accuracy is calculated based on the upper two quadrants and is used to evaluate the IVMs of MLLMs. Hallucinations, on the other hand, are assessed based on the lower two quadrants.
    }
    \label{x_fig:9}
\end{figure}

When evaluating the IVMs level of MLLMs, we use the attention accuracy metric.
This metric is calculated based on samples where the model has already provided correct answers.
In fact, by combining answer correctness with \cref{def:attn_correct}, we can define four quadrants, as illustrated in \cref{x_fig:9}.
Therefore, we can calculate the attention accuracy in cases where the model’s answers are incorrect to assess the level of EVMs.
For some general visual understanding tasks (such as Document VQA \cite{DocVQA} and Textbook QA \cite{TQA}), EVMs in MLLMs typically manifest as hallucinations.

In the STME benchmark, four hard tasks are selected: Document VQA, Text-Rich Images QA \cite{SlideVQA}, Textbook QA, and Space Understanding \cite{nuscenes}.
Attention accuracy is then calculated in cases where the model’s answers are incorrect.
As shown in \cref{x_tab:Hallucinations}, attention accuracy aligns with existing hallucination benchmarks, suggesting that hallucinations can indeed be significantly reduced as the model size increases.
\begin{table}
  \centering
  \resizebox{0.7\linewidth}{!}{
  \begin{tabular}{@{}lccccc@{}}
    \toprule
    & Params & \textbf{Attn Acc (\%)} & HallusionBench & POPE\\
    \midrule
    Qwen2VL \cite{qwen2vl} & 2B & 58.2 & 42.4 & 87.3 \\
    Qwen2VL \cite{qwen2vl} & 7B & 88.5 & 50.4 & 88.4\vspace{3px} \\
    InternVL2 \cite{internvl2} & 1B & 45.5 & 34.3 & 84.9 \\
    InternVL2 \cite{internvl2} & 2B & 56.3 & 38.0 & 85.2 \\
    InternVL2 \cite{internvl2} & 4B & 62.7 & 42.4 & 84.6 \\
    InternVL2 \cite{internvl2} & 8B & 81.8 & 45.0 & 84.2 \\
    InternVL2 \cite{internvl2} & 26B & 85.3 & 51.5 & 86.4\vspace{3px} \\
    LLaVA-OV \cite{llava-ov} & 0.5B & 71.4 & 27.9 & 87.8 \\
    LLaVA-OV \cite{llava-ov} & 7B & 80.6 & 31.6 & 88.4 \\
    \bottomrule
  \end{tabular}
  }
  \caption{
  Compared to the hallucination benchmarks HallusionBench \cite{HallusionBench} and POPE \cite{POPE}, our evaluation method demonstrates consistency, suggesting that attention accuracy can also be used to assess the hallucination level of MLLMs.
  }
  \label{x_tab:Hallucinations}
\end{table}

\subsection{Experiments for Sufficiency Proof}
The experiments in \cref{sec:experiments} positively validate the effectiveness of attention accuracy in assessing IVMs, that is, when the model exhibits IVMs, attention accuracy decreases accordingly.
However, in OCR tasks, while MLLMs can successfully attend to the target image, they may fail to provide the correct answer due to limitations in their fine-grained visual capabilities.
For example, in the example of EVMs shown in \cref{fig:1_examples}, the model may correctly locate the image containing the relevant digits but struggle to accurately recognize all the numbers due to insufficient OCR capabilities.

To analyze this, we separately examine the ``Image Needle in a Haystack'' \cite{Needle} task.
This task presents multiple images, with only one containing a string of special digits.
The models‘ objective is to locate that string among the images.
In such cases, the models typically demonstrate the ability to identify the image containing the digits but struggle to fully and accurately recognize the entire string due to limited OCR capability.
Therefore, we select the sample that meets this condition to calculate attention accuracy of models.

The final results in \cref{x_tab:ocr_result} show that, for the Qwen2VL series models, InternVL2 models ranging from 2B to 26B, and the LLaVA-OneVision-7B model, a consistent conclusion emerges:
in samples where OCR recognition is correct, or where the model's output contains a string of digits but OCR limitations lead to inaccuracies, the model's attention distribution converges to the target image.
This further supports the idea that the image to which attention converges is the one the model ultimately focuses on, and thus, attention accuracy can be used to relatively accurately assess IVMs in MLLMs.
\begin{table}
  \centering
  \resizebox{0.48\linewidth}{!}{
  \begin{tabular}{@{}lccc@{}}
    \toprule
    & Params & \textbf{Attn Acc (\%)}\\
    \midrule
    Qwen2VL \cite{qwen2vl} & 2B & 99.3 \\
    Qwen2VL \cite{qwen2vl} & 7B & 100\vspace{3px} \\
    InternVL2 \cite{internvl2} & 1B & 90.2 \\
    InternVL2 \cite{internvl2} & 2B & 99.0 \\
    InternVL2 \cite{internvl2} & 4B & 99.3 \\
    InternVL2 \cite{internvl2} & 8B & 100 \\
    InternVL2 \cite{internvl2} & 26B & 99.3\vspace{3px} \\
    LLaVA-OV \cite{llava-ov} & 0.5B & 83.6 \\
    LLaVA-OV \cite{llava-ov} & 7B & 100 \\
    \bottomrule
  \end{tabular}
  }
  \caption{
  Compared to the hallucination benchmarks HallusionBench \cite{HallusionBench} and POPE \cite{POPE}, our evaluation method demonstrates consistency, suggesting that attention accuracy can also be used to assess the hallucination level of MLLMs.
  }
  \label{x_tab:ocr_result}
\end{table}

\newpage
\section{Examples of STME and Attention Distribution in MLLMs}
We present several examples from the STME benchmark, accompanied by attention distribution heatmaps that illustrate the changes during inference on these tasks using the MLLMs.
\label{x_subsec:more_tendency}
\begin{figure}[!htbp]
    \centering
    \begin{subfigure}{0.46\linewidth}
        \centering
        \includegraphics[width=\linewidth]{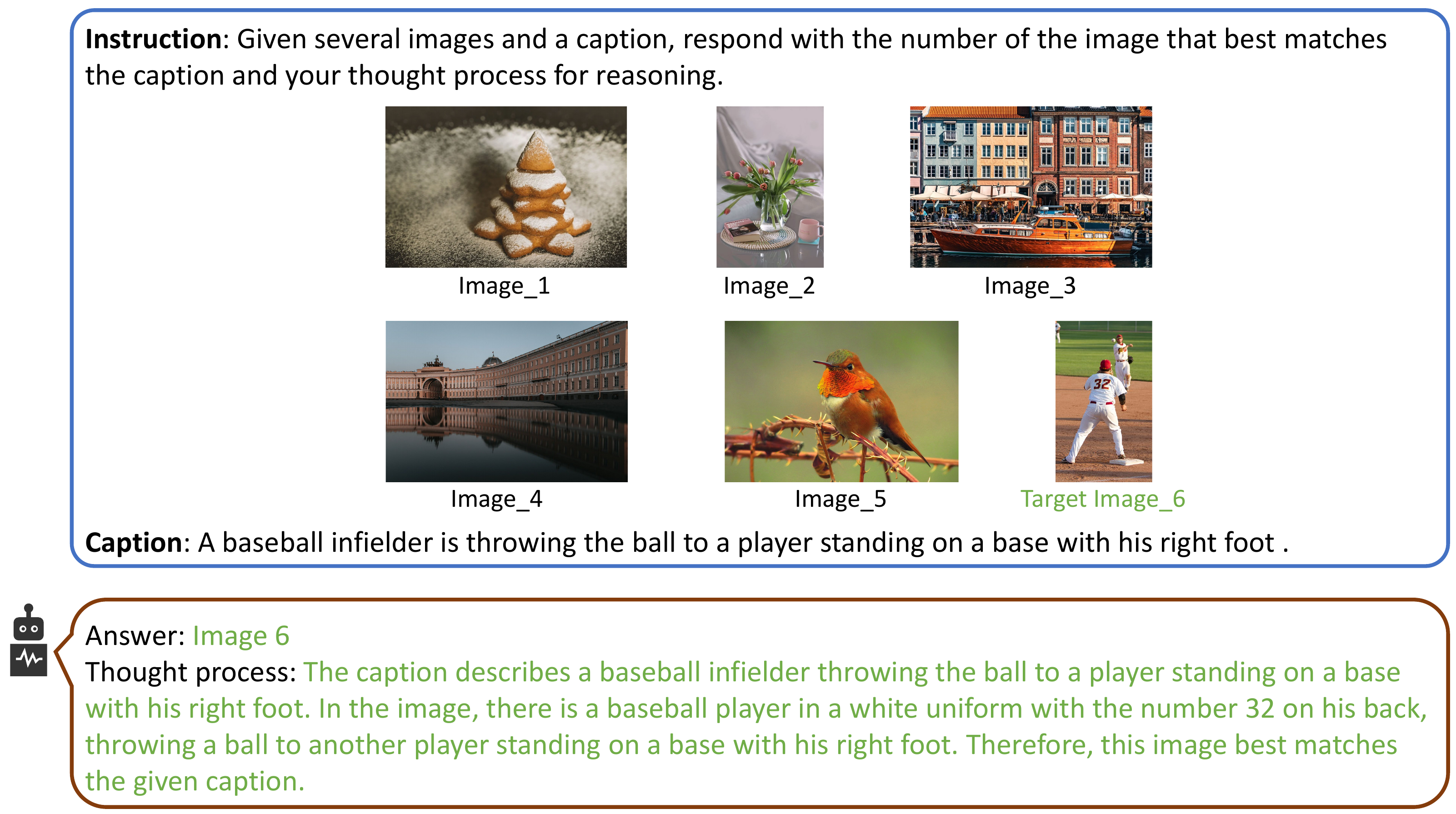}
    \end{subfigure}
    \begin{subfigure}{0.52\linewidth}
        \centering
        \includegraphics[width=\linewidth]{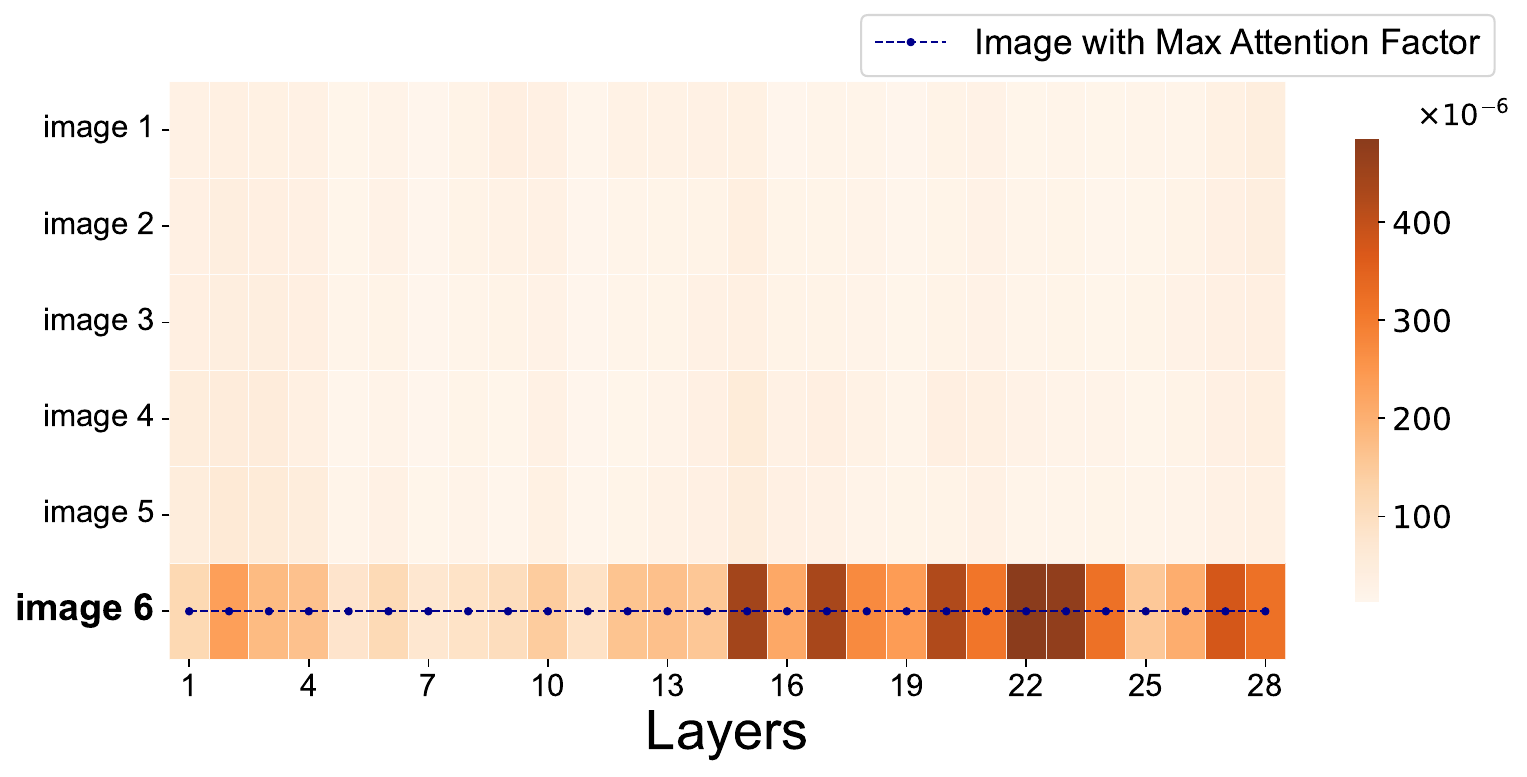}
    \end{subfigure}
    \caption{
    This demonstrates Qwen2VL-2B performing a \textbf{caption matching} task, with the sixth image serving as the target image. The model correctly identifies the answer, and its attention appropriately converges to the correct image.
    }
\end{figure}
\begin{figure}[!htbp]
    \centering
    \begin{subfigure}{0.46\linewidth}
        \centering
        \includegraphics[width=\linewidth]{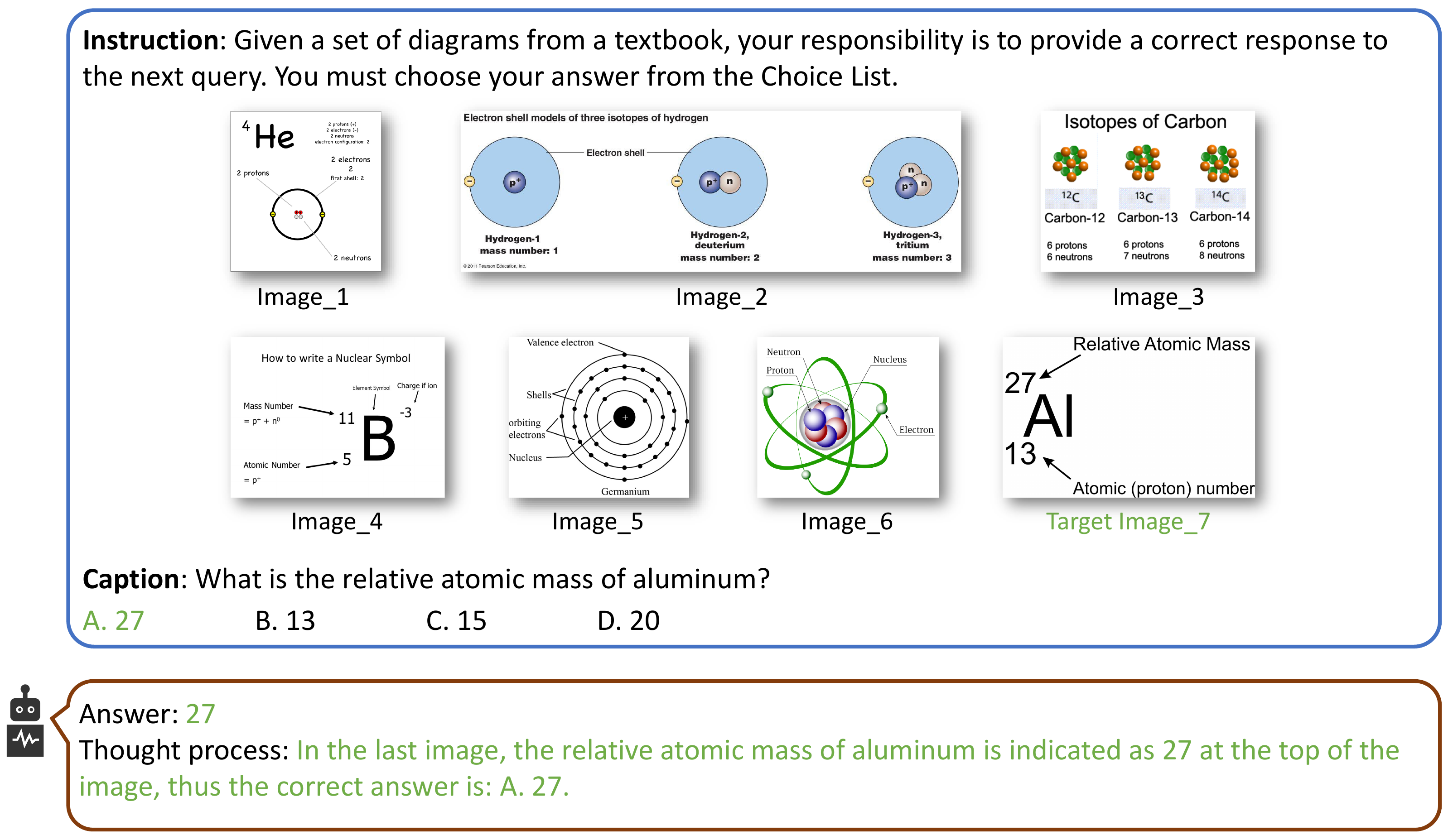}
    \end{subfigure}
    \begin{subfigure}{0.52\linewidth}
        \centering
        \includegraphics[width=\linewidth]{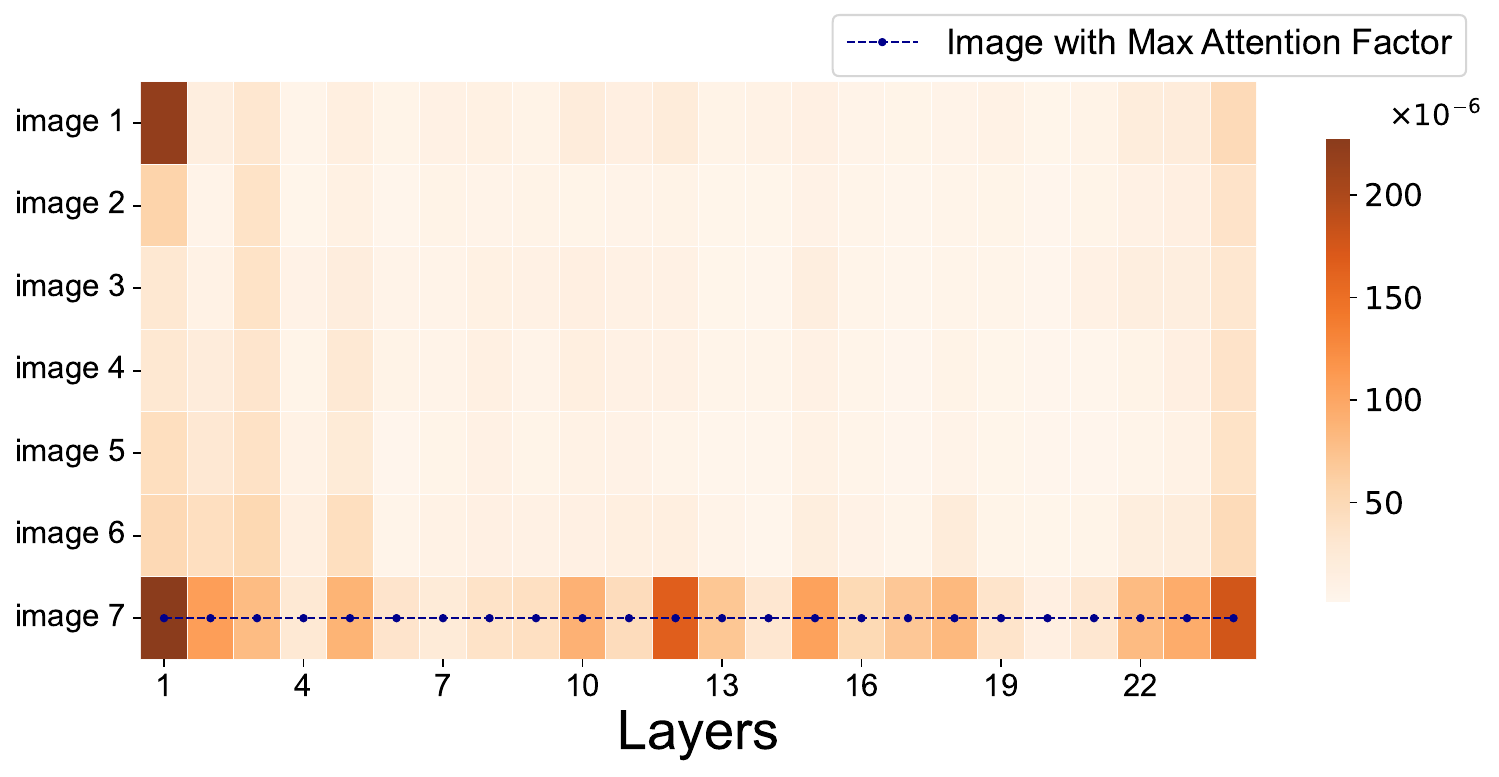}
    \end{subfigure}
    \caption{
    This demonstrates LLaVA-OneVision-0.5B performing a \textbf{Textbook QA} task, with the seventh image as the target image. The model correctly identified the relative atomic mass of aluminum and provided the correct answer with a reasonable explanation. The attention distribution shows that the model focused on the target image (We have posed the question in a text-based format to ensure the model has to fully understand the image to answer correctly).
    }
\end{figure}
\begin{figure}[!htbp]
    \centering
    \begin{subfigure}{0.46\linewidth}
        \centering
        \includegraphics[width=\linewidth]{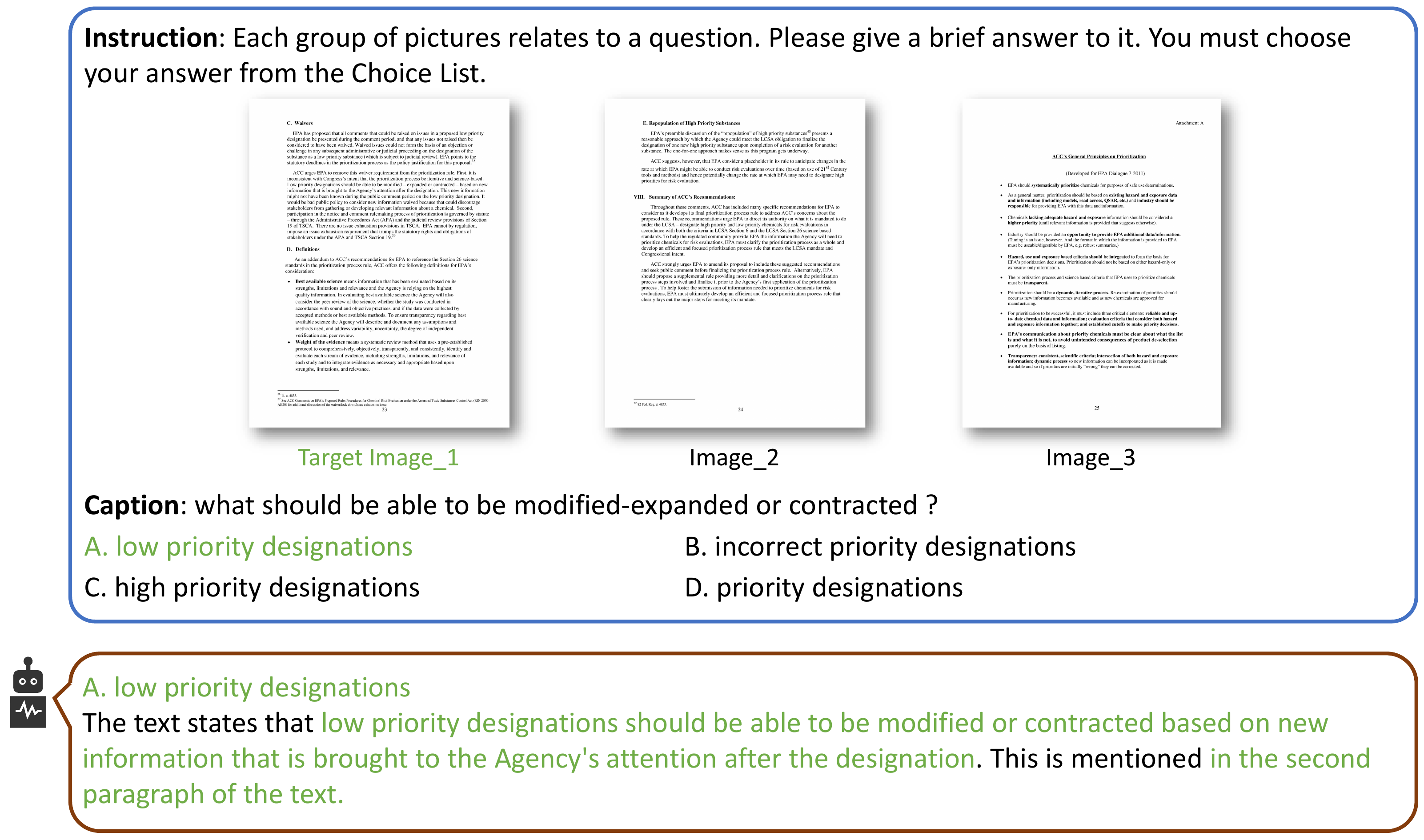}
    \end{subfigure}
    \begin{subfigure}{0.52\linewidth}
        \centering
        \includegraphics[width=\linewidth]{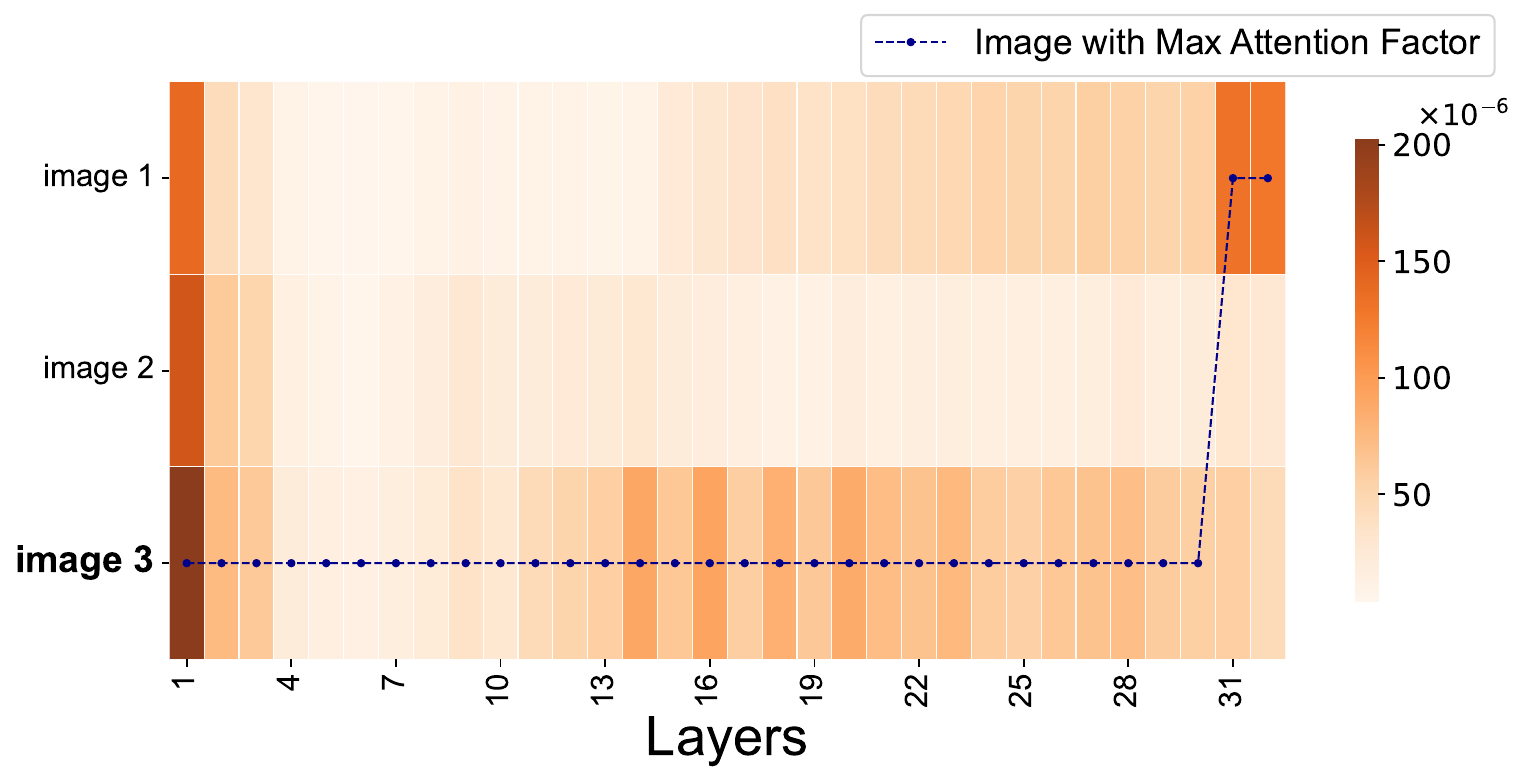}
    \end{subfigure}
    \caption{
    InternVL2-4B performs a \textbf{Document VQ} task with the first image as the target. After processing the text, the model extracts relevant information from the image and provides the correct answer. An interesting pattern appears in the attention heatmap: the model focuses on the last image in earlier layers and shifts attention to the target image only in the final two layers. This suggests that handling large amounts of text requires multiple layers to fully process the information in the target image.
    }
\end{figure}
\begin{figure}[htbp]
    \centering
    \begin{subfigure}{\linewidth}
        \centering
        \includegraphics[width=\linewidth]{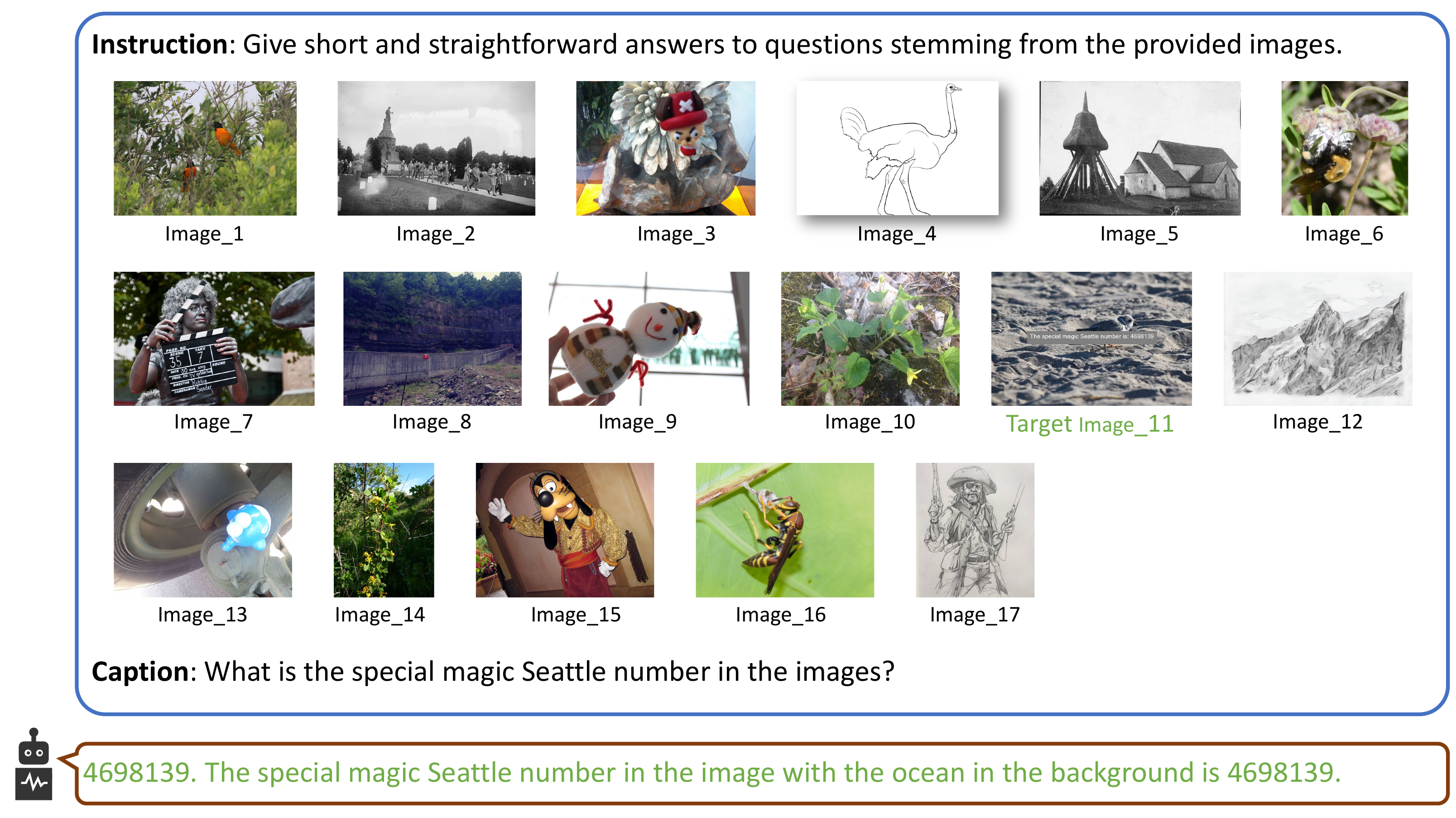}
        \vspace{15px}
    \end{subfigure} 

    \begin{subfigure}{\linewidth}
        \centering
        \includegraphics[width=0.95\linewidth]{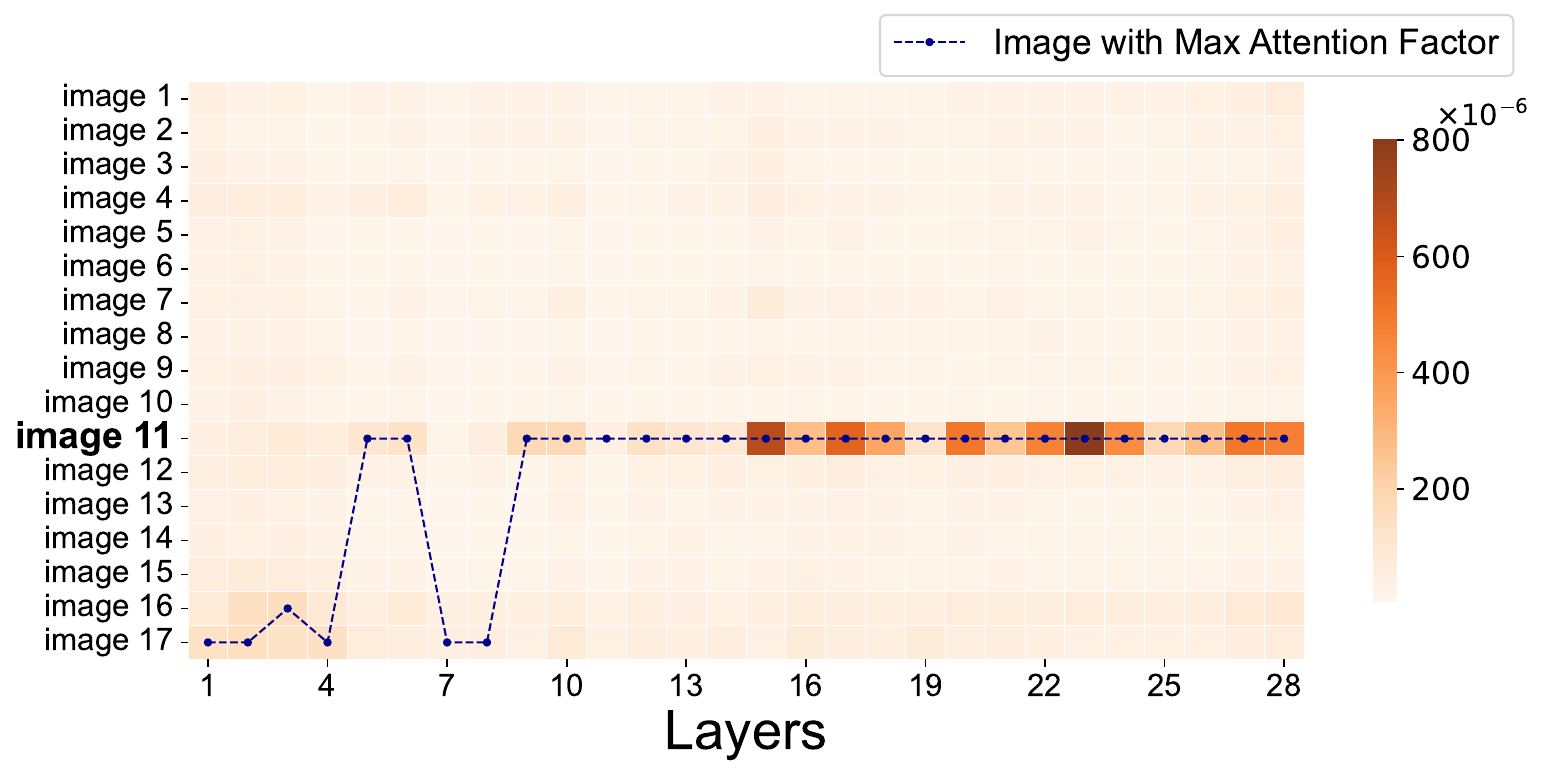}
    \end{subfigure}
    \caption{
    Qwen2VL-7B performs a \textbf{Image Needle in a Haystack} task with the eleventh image as the target image. The model successfully detected the special digits s in the target image from a set of 17 images using only 8 layers and accurately recognized the result. This demonstrates the strong performance of the Qwen2VL's visual encoder.
    }
\end{figure}

\newpage
\section{Sensitivity Analysis of Attention Accuracy}
\label{x_sec:other_model_metric}
On the STME benchmark, we analyze the sensitivity of the attention accuracy metric across models of different series and parameter scales.
As $N$ increases, the computed attention accuracy exhibits systematic fluctuations.
\begin{itemize}
    \item Overall, the LND metric achieves the best performance, but it also exhibits the highest volatility as $N$ increases.
    \item As $N$ increases, the curves for the three metrics generally show an increasing trend followed by a decrease.
    \item Models with smaller parameter scales reach the inflection point more quickly, particularly InternVL2-2B and LLaVA-OneVision-0.5B. This suggests that smaller models contain relatively less visual information, requiring fewer layers for alignment and interpretation. This indirectly supports the notion that smaller models have a lower performance ceiling compared to larger models.
    \item When performing inference on easy tasks, the curves corresponding to the three metrics reach the inflection point more quickly. This indicates that the model converges faster on simpler tasks and slower on more challenging ones.
\end{itemize}
\begin{figure}[!htbp]
    \centering
    \begin{subfigure}{0.75\linewidth}
        \centering
        \includegraphics[width=\linewidth]{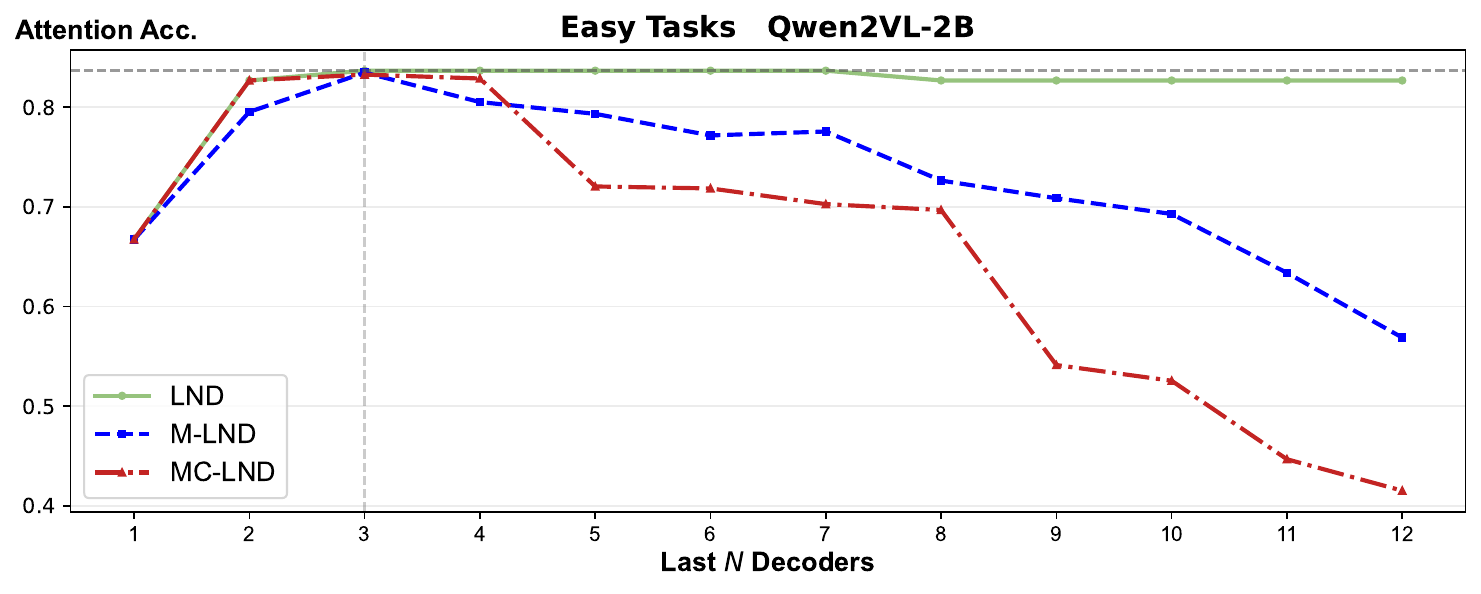}
    \end{subfigure}
    \begin{subfigure}{0.75\linewidth}
        \centering
        \includegraphics[width=\linewidth]{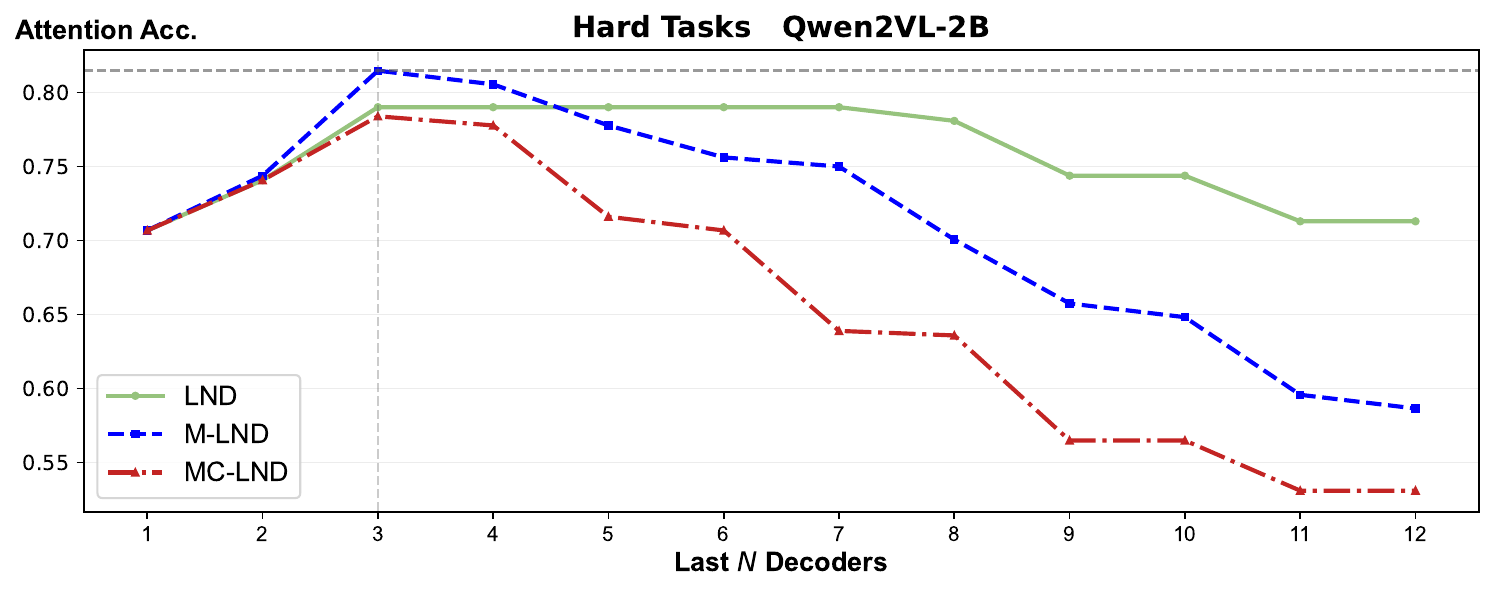}
    \end{subfigure}
    \vspace{-5px}
    \caption{
    The attention accuracy of Qwen2VL-2B shows that, as $N$ increases, the curves for M-LND and MC-LND quickly decline, while the LND method remains more robust.
    On both easy and hard tasks, all three metrics reach their inflection points at approximately $N=3$, with the variation trends being relatively similar.
    }
\end{figure}
\begin{figure}[!htbp]
    \centering
    \begin{subfigure}{0.75\linewidth}
        \centering
        \includegraphics[width=\linewidth]{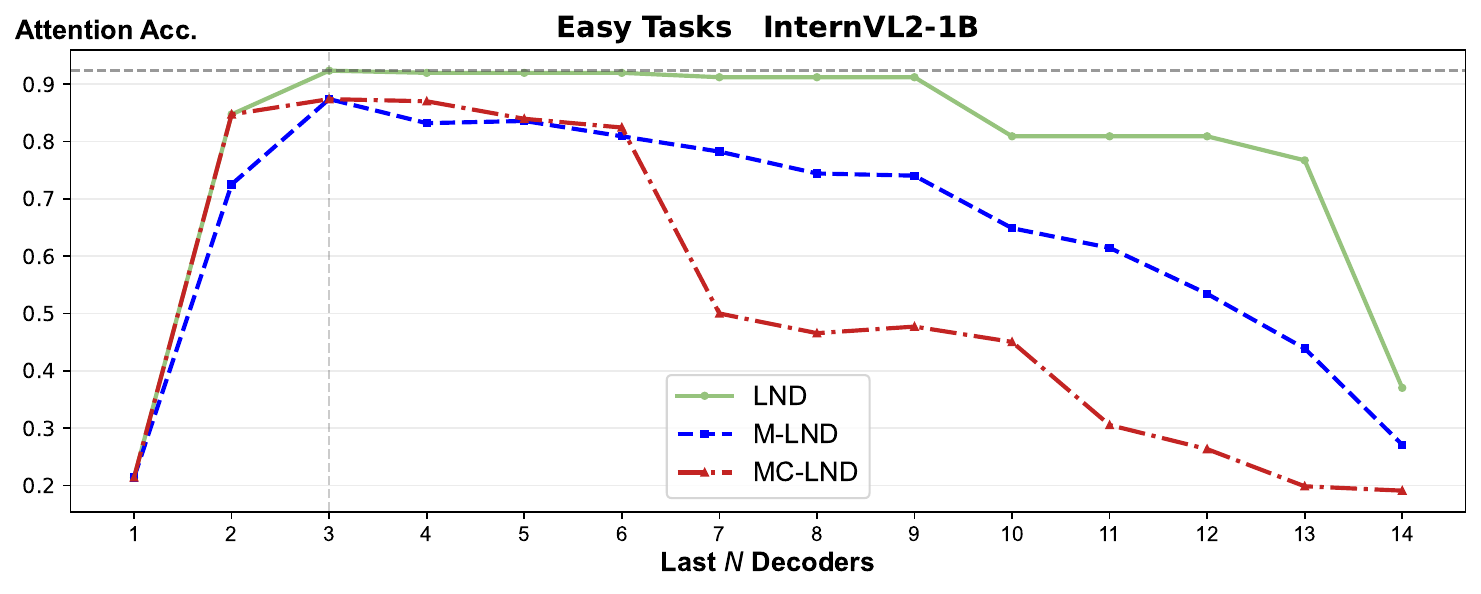}
    \end{subfigure}
    \begin{subfigure}{0.75\linewidth}
        \centering
        \includegraphics[width=\linewidth]{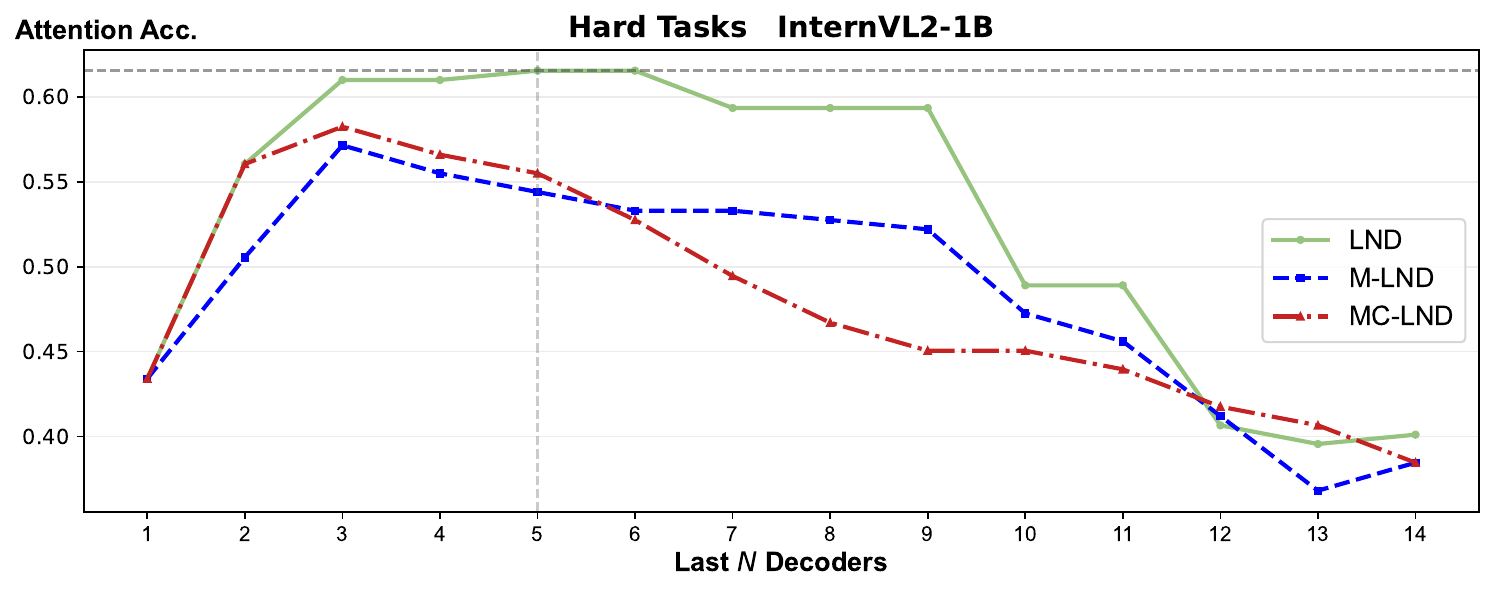}
    \end{subfigure}
    \vspace{-5px}
    \caption{
    The attention accuracy of InternVL2-1B, calculated using three metrics, on the easy and hard datasets.
    }
\end{figure}
\begin{figure}[!htbp]
    \centering
    \begin{subfigure}{0.75\linewidth}
        \centering
           \includegraphics[width=\linewidth]{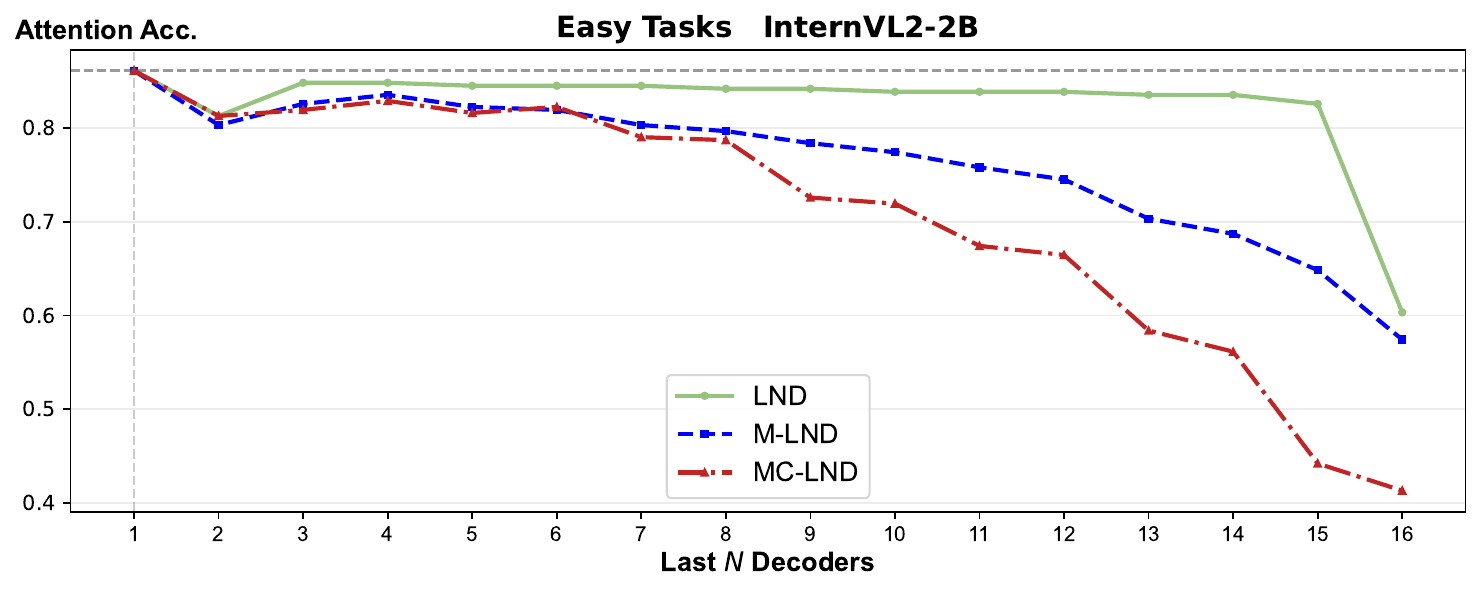}
    \end{subfigure}
    \begin{subfigure}{0.75\linewidth}
        \centering
        \includegraphics[width=\linewidth]{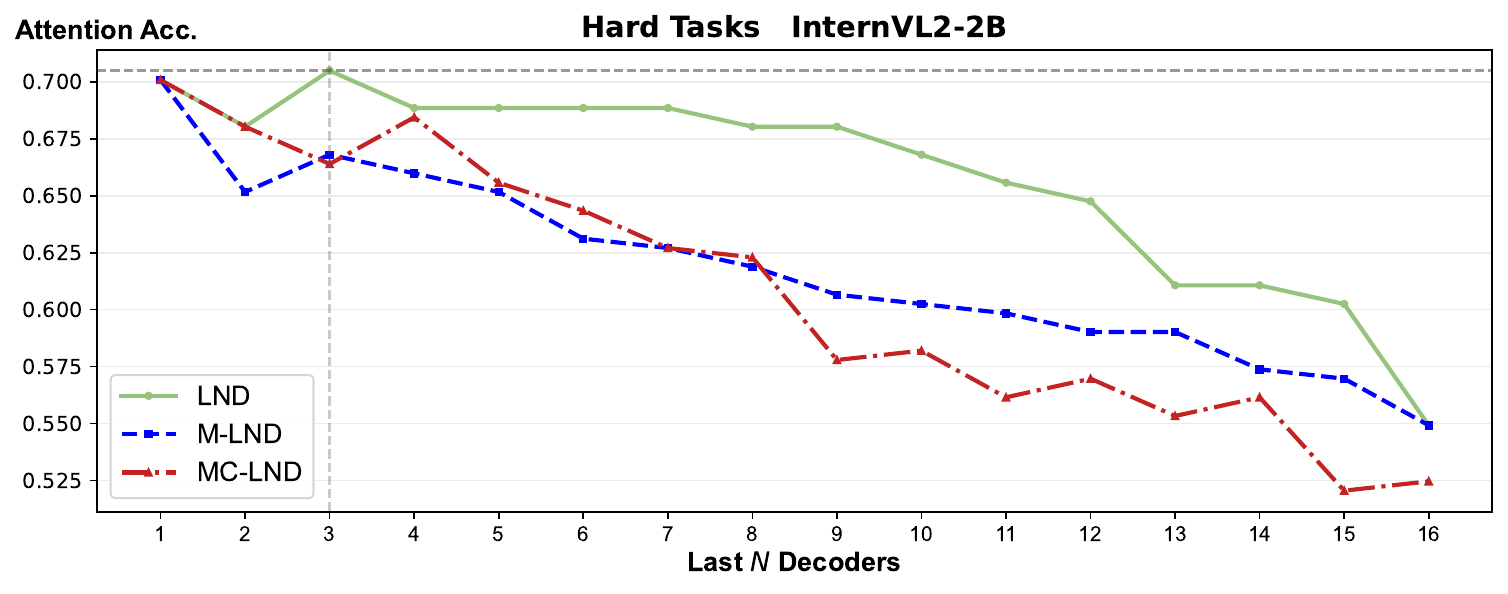}
    \end{subfigure}
    \vspace{-5px}
    \caption{
    The attention accuracy of InternVL2-2B, calculated using three metrics, on the easy and hard datasets.
    }
\end{figure}
\begin{figure}[!htbp]
    \centering
    \begin{subfigure}{0.75\linewidth}
        \centering
        \includegraphics[width=\linewidth]{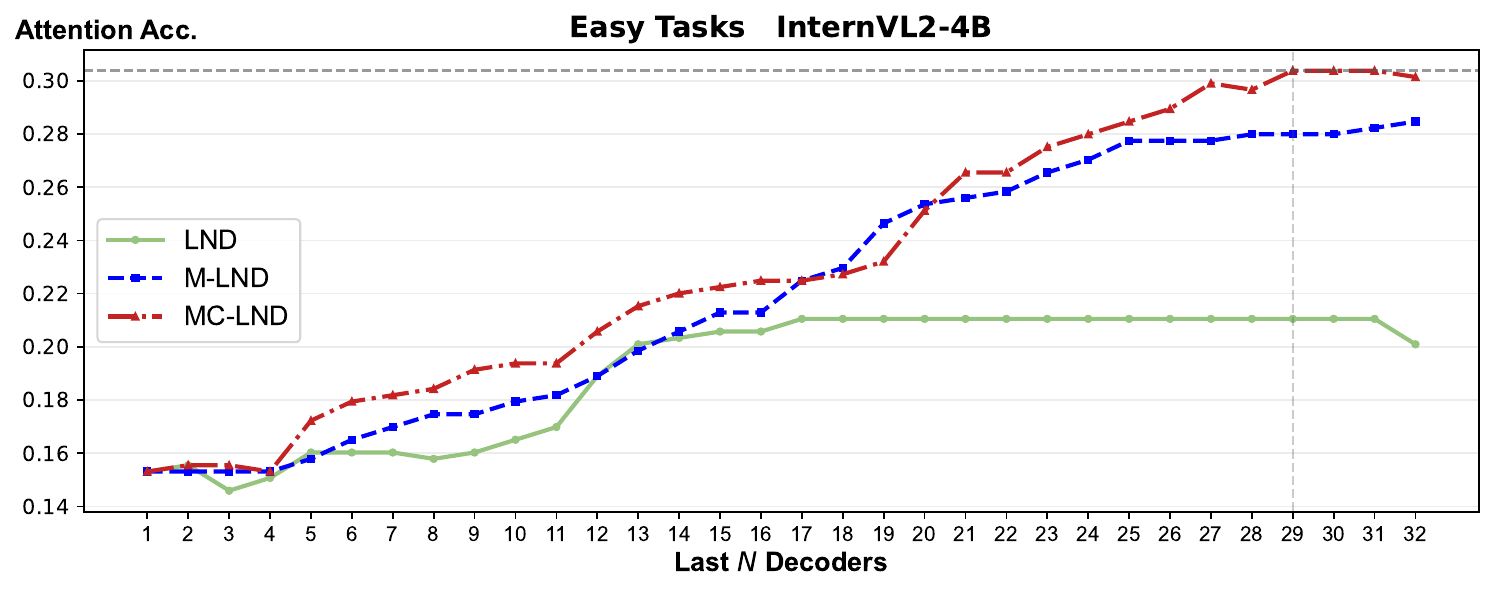}
    \end{subfigure}
    \begin{subfigure}{0.75\linewidth}
        \centering
        \includegraphics[width=\linewidth]{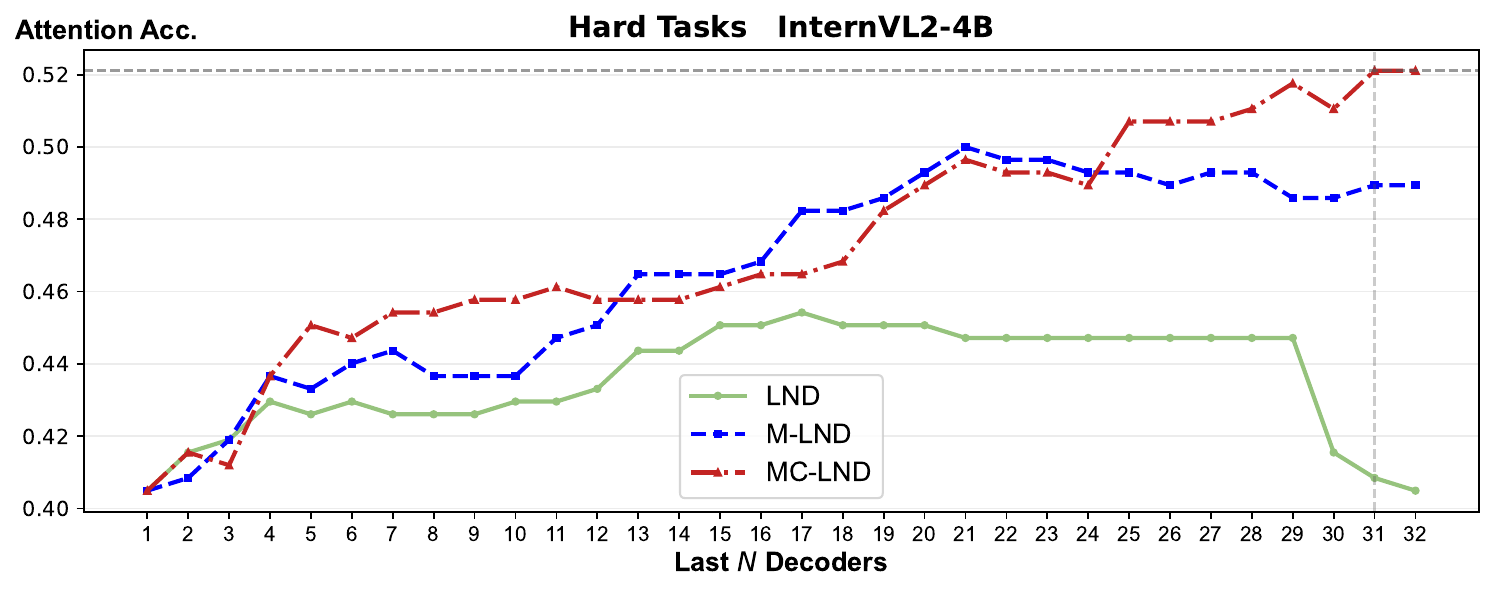}
    \end{subfigure}
    \vspace{-5px}
    \caption{
    The attention accuracy of InternVL2-4B, calculated using three metrics, on the easy and hard datasets.
    }
\end{figure}
\begin{figure}[!htbp]
    \centering
    \begin{subfigure}{0.75\linewidth}
        \centering
        \includegraphics[width=\linewidth]{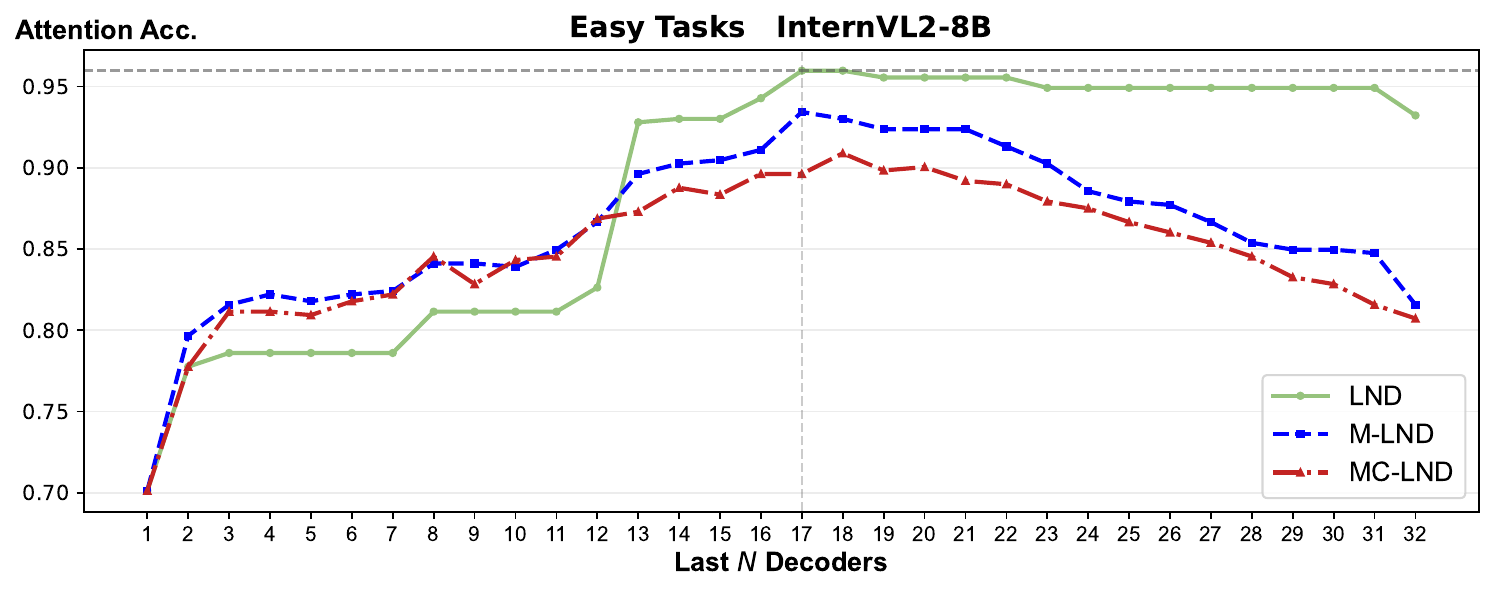}
    \end{subfigure}
    \begin{subfigure}{0.75\linewidth}
        \centering
        \includegraphics[width=\linewidth]{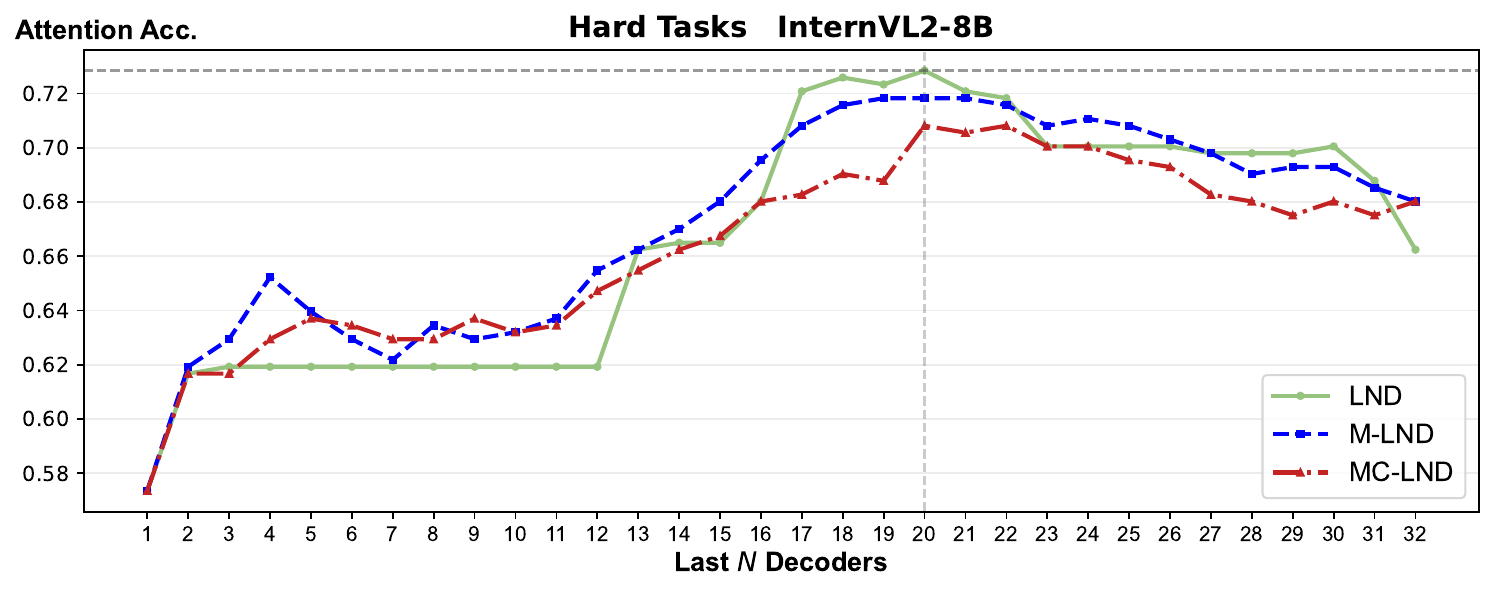}
    \end{subfigure}
    \vspace{-5px}
    \caption{
    The attention accuracy of InternVL2-8B, calculated using three metrics, on the easy and hard datasets.
    }
\end{figure}
\begin{figure}[!htbp]
    \centering
    \begin{subfigure}{0.75\linewidth}
        \centering
        \includegraphics[width=\linewidth]{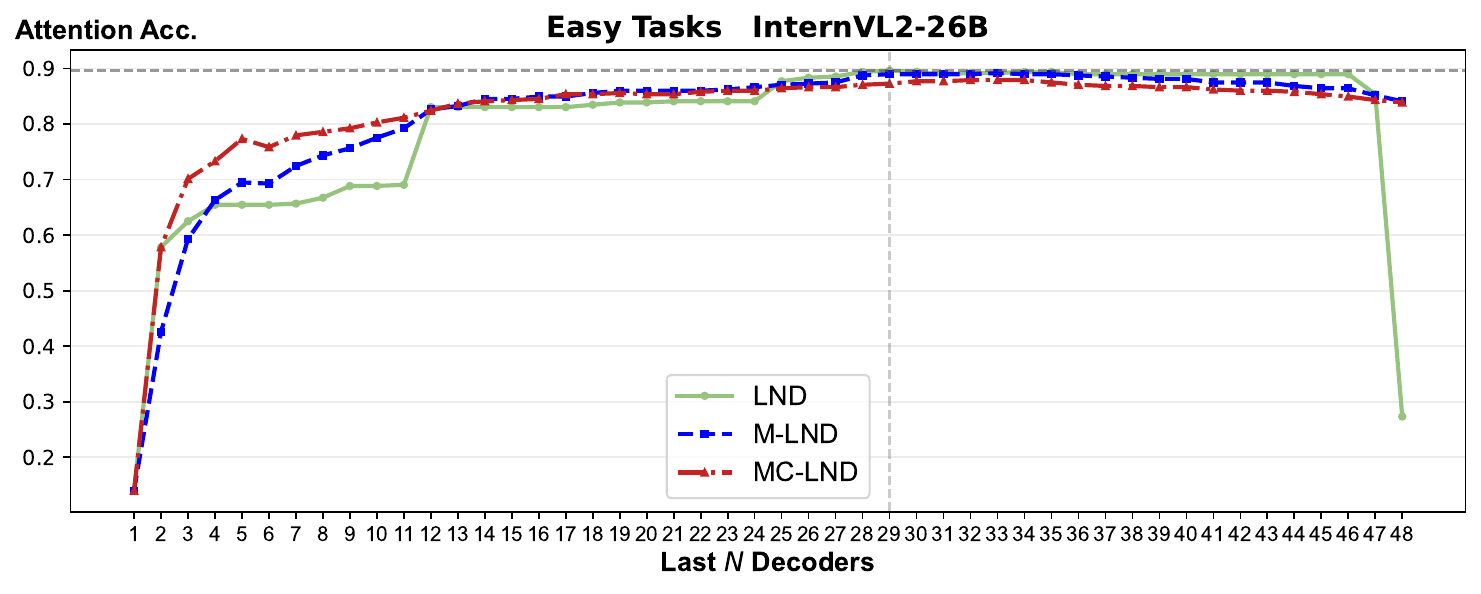}
    \end{subfigure}
    \begin{subfigure}{0.75\linewidth}
        \centering
        \includegraphics[width=\linewidth]{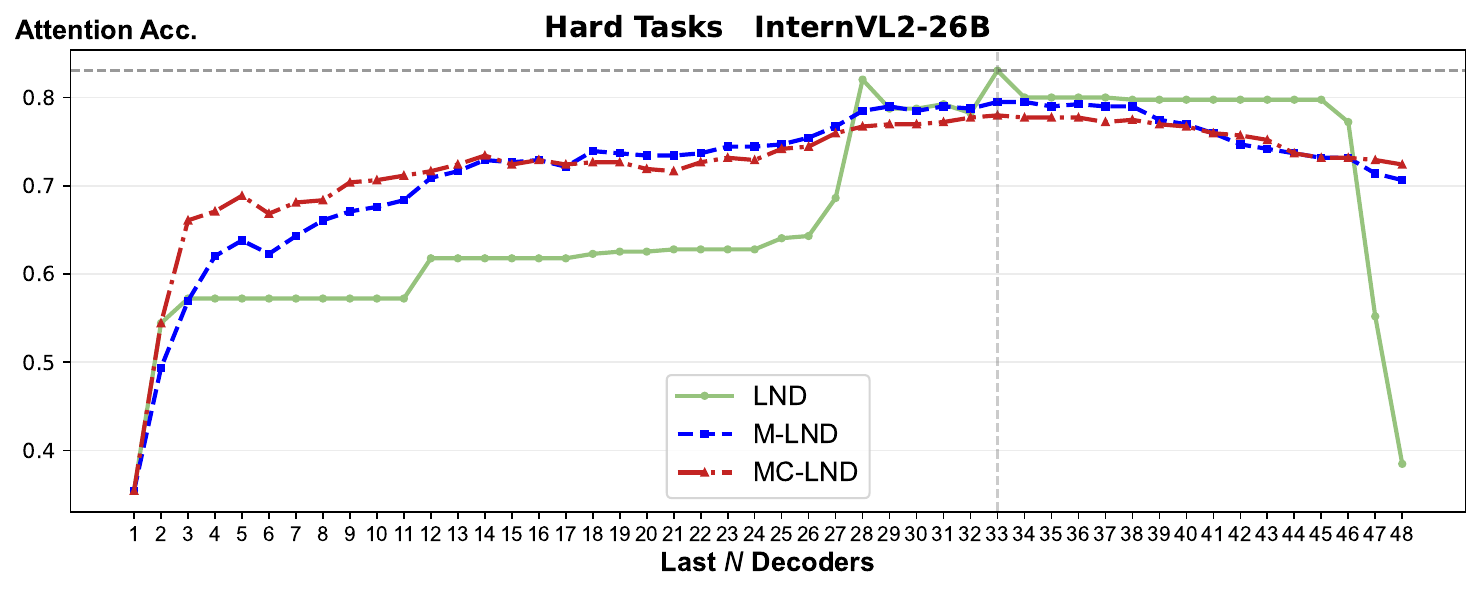}
    \end{subfigure}
    \vspace{-5px}
    \caption{
    The attention accuracy of InternVL2-26B, calculated using three metrics, on the easy and hard datasets.
    }
\end{figure}
\begin{figure}[!htbp]
    \centering
    \begin{subfigure}{0.75\linewidth}
        \centering
        \includegraphics[width=\linewidth]{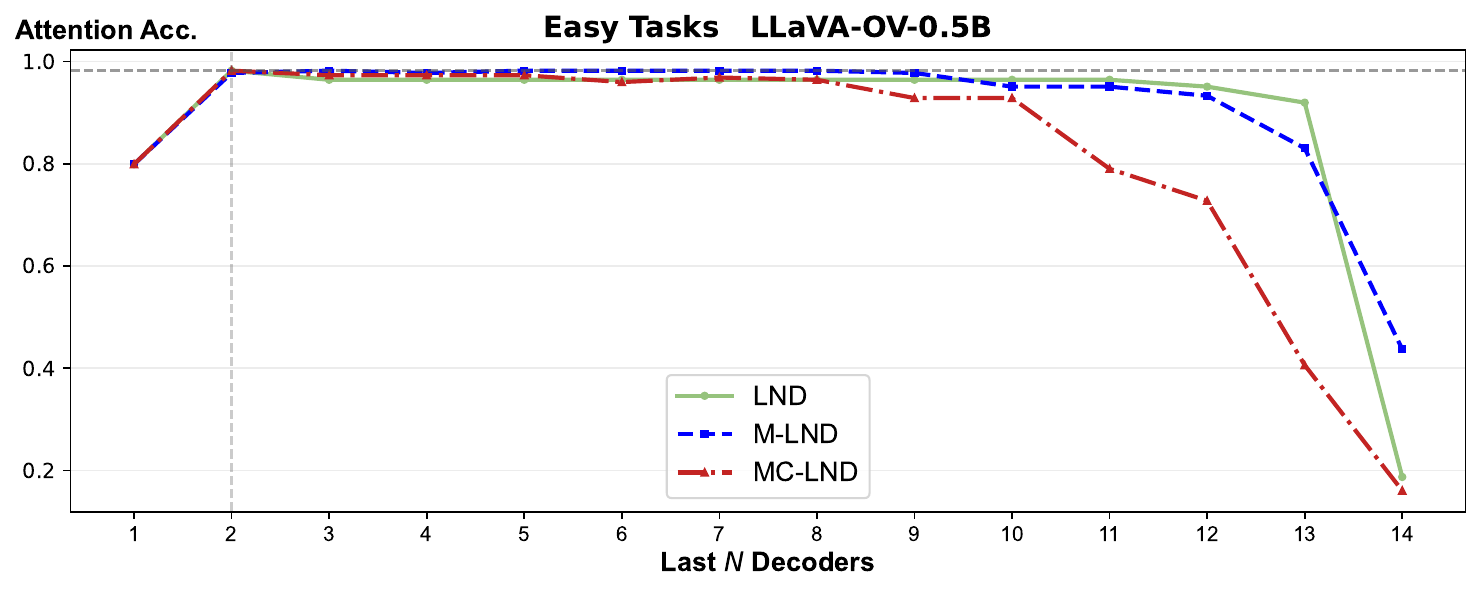}
    \end{subfigure}
    \begin{subfigure}{0.75\linewidth}
        \centering
        \includegraphics[width=\linewidth]{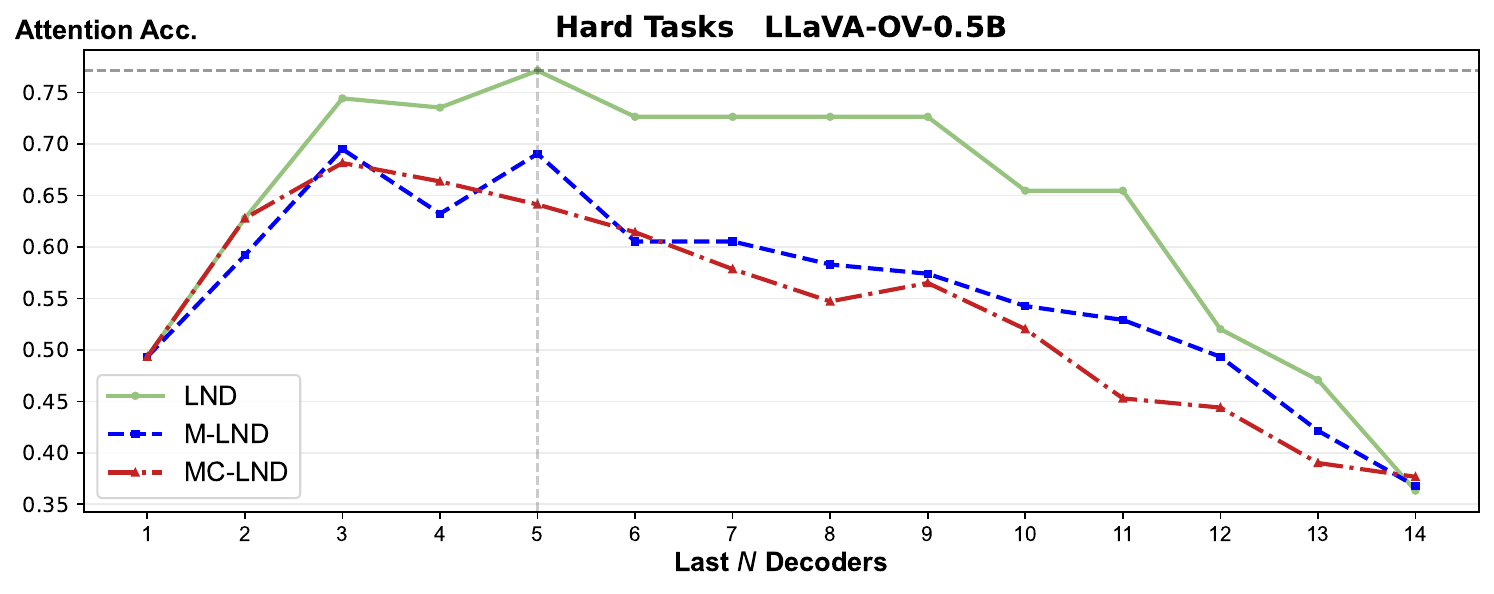}
    \end{subfigure}
    \vspace{-5px}
    \caption{
    The attention accuracy of LLaVA-OneVision-0.5B, calculated using three metrics, on the easy and hard datasets.
    }
\end{figure}
\begin{figure}[!htbp]
    \centering
    \begin{subfigure}{0.75\linewidth}
        \centering
        \includegraphics[width=\linewidth]{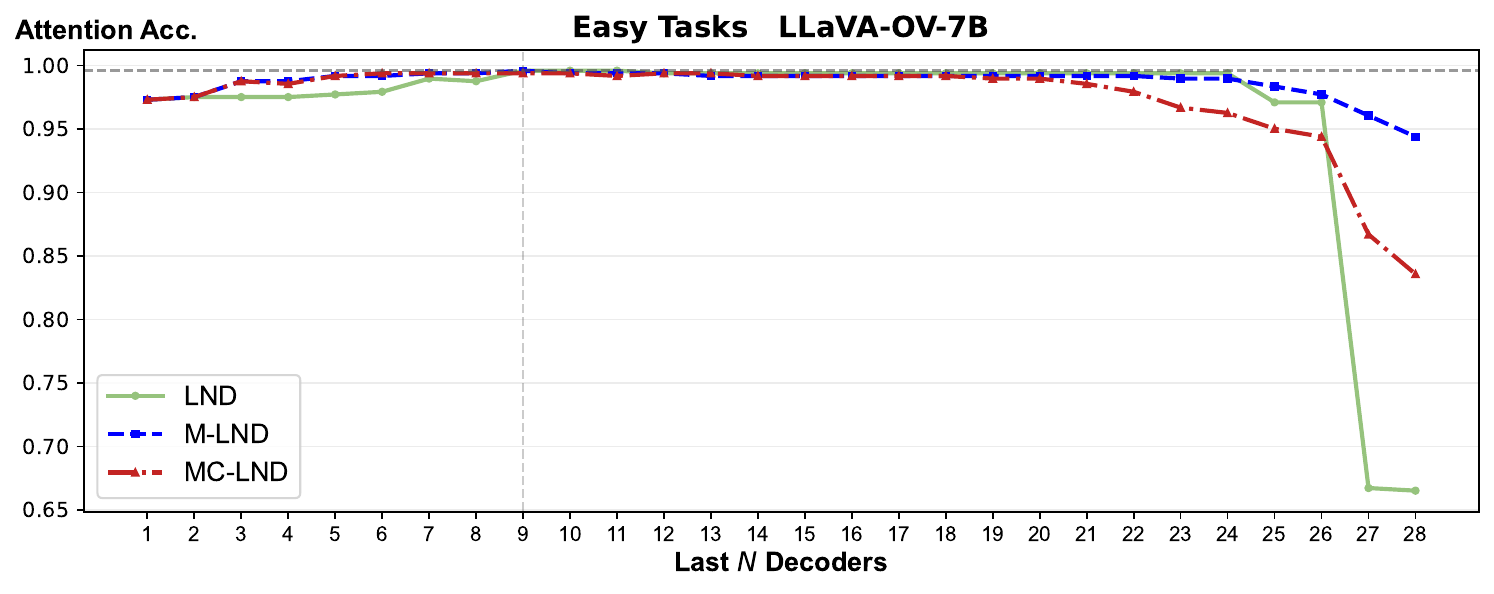}
    \end{subfigure}
    \begin{subfigure}{0.75\linewidth}
        \centering
        \includegraphics[width=\linewidth]{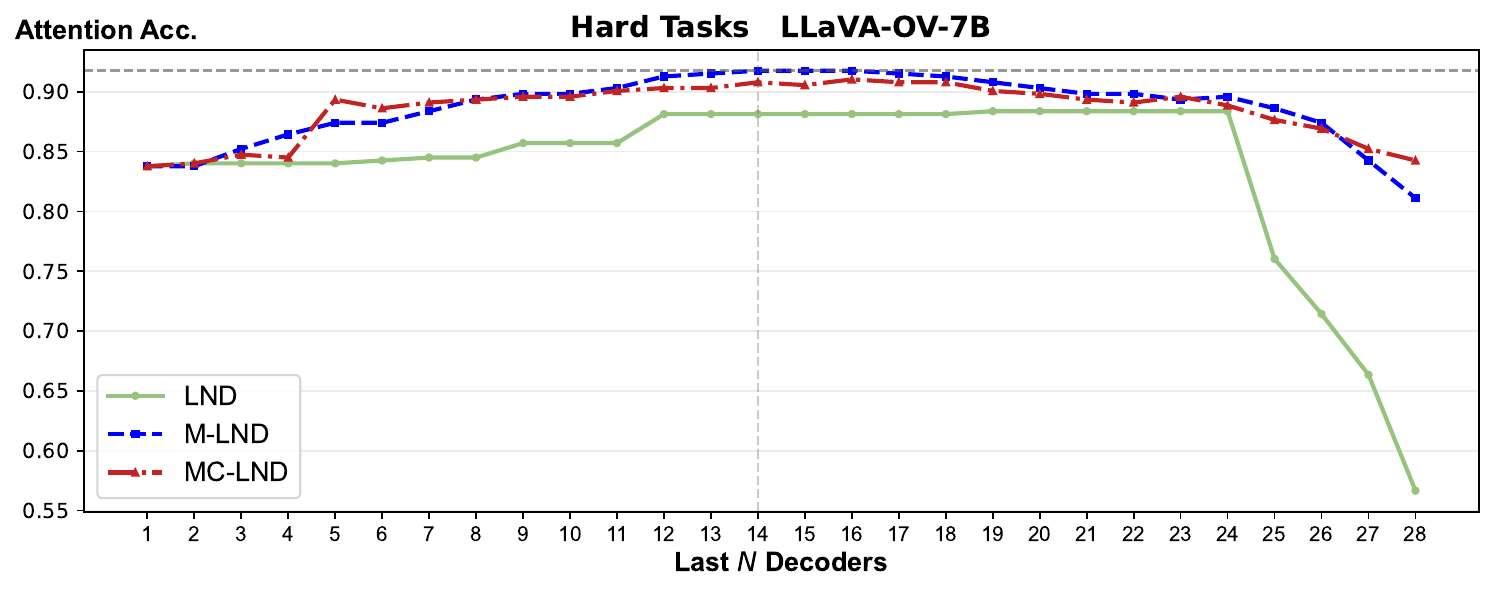}
    \end{subfigure}
    \caption{
    The attention accuracy of LLaVA-OneVision-7B, calculated using three metrics, on the easy and hard datasets.
    }
\end{figure}

\end{document}